\documentclass{article}
\usepackage{amssymb}
\usepackage{ifthen}
\usepackage{hyperref}
\usepackage{cite}
\usepackage{amsmath}
\usepackage{amsmath,amsfonts}
\usepackage{graphicx}
\usepackage{mathtools}
\usepackage{overpic}
\usepackage{comment}
\usepackage{booktabs}
\usepackage{multirow}
\usepackage{duckuments}
\usepackage{svg}
\usepackage{subcaption}

\usepackage[preprint]{corl_2025}

\title{Steerable Scene Generation\\with Post Training and Inference-Time Search}

\author{
  Nicholas Pfaff$^1$, Hongkai Dai$^2$, Sergey Zakharov$^2$, Shun Iwase$^{2,3}$, Russ Tedrake$^{1,2}$ \\
  $^1$Massachusetts Institute of Technology, $^2$Toyota Research Institute, $^3$Carnegie Mellon University
}

\begin{document}
\maketitle

\begin{figure}[!htb]
\centerline{
\includegraphics[width=\linewidth, keepaspectratio]
{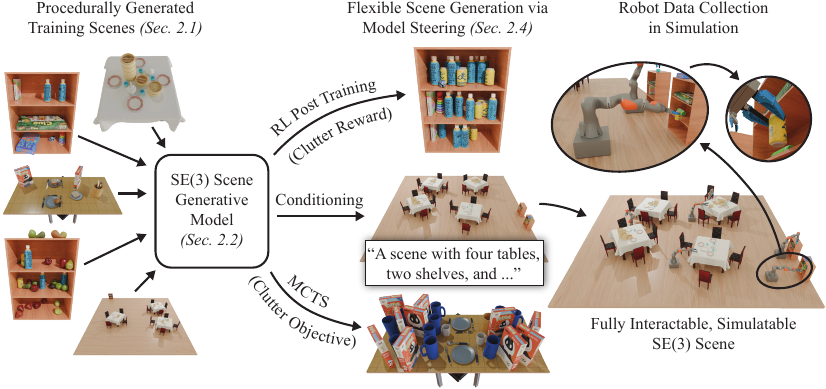}
}
\caption{\textbf{Overview of our approach.} 
We train a diffusion-based generative model on SE(3) scenes generated by procedural models, then adapt it to downstream objectives via reinforcement learning-based post training, conditional generation, or inference-time search. The resulting scenes are physically feasible and fully interactable. We demonstrate teleoperated interaction in a subset of generated scenes using a mobile KUKA iiwa robot.
}
\label{fig:teaser}
\end{figure}

\begin{abstract}

Training robots in simulation requires diverse 3D scenes that reflect the specific challenges of downstream tasks. However, scenes that satisfy strict task requirements, such as high-clutter environments with plausible spatial arrangement, are rare and costly to curate manually. Instead, we generate large-scale scene data using procedural models that approximate realistic environments for robotic manipulation, and adapt it to task-specific goals.
We do this by training a unified diffusion-based generative model that predicts which objects to place from a fixed asset library, along with their SE(3) poses. This model serves as a flexible scene prior that can be adapted using reinforcement learning-based post training, conditional generation, or inference-time search, steering generation toward downstream objectives even when they differ from the original data distribution.
Our method enables goal-directed scene synthesis that respects physical feasibility and scales across scene types. We introduce a novel MCTS-based inference-time search strategy for diffusion models, enforce feasibility via projection and simulation, and release a dataset of over 44 million SE(3) scenes spanning five diverse environments.
Website with videos, code, data, and model weights: \url{https://steerable-scene-generation.github.io/}

\end{abstract}


\section{Introduction}
\label{sec:introduction}

Robots increasingly rely on data-intensive learning methods, making simulation a promising strategy for scalable training and evaluation \cite{on_the_use_of_sim_in_robotics_2021, sim2real_survey_2020, wei2025empiricalanalysissimandrealcotraining, maddukuri2025simandrealcotrainingsimplerecipe, eppner2020acronymlargescalegraspdataset}. As robotics shifts toward foundation models, it is encouraging that the demand for large and diverse training datasets will only increase \cite{foundation_models_in_robotics, embodimentcollaboration2024openxembodimentroboticlearning, khazatsky2025droidlargescaleinthewildrobot}.
However, acquiring scenes that meaningfully challenge robot capabilities or reflect human teleoperator preferences remains difficult, as such scenes are rare, expensive to curate, and task-specific. For example, a robot may need to operate in highly cluttered environments or interact with specific object categories. Instead of manually authoring such scenes, we propose training a unified generative model on large-scale procedurally generated data and adapting it to downstream objectives using reinforcement learning-based post training, conditional generation, and inference-time search.
\\
Recent work has advanced automatic scene creation at both the object \cite{google_scanned_objects, pfaff2025_scalable_real2sim} and scene level \cite{physcene, robocasa2024}. We focus on the latter, where the task is to select objects from a fixed library and place them at continuous SE(3) poses.
Classical approaches to scene synthesis rely on procedural modeling, where object relationships are encoded as rule sets or grammars \cite{furniture_layout_using_interior_design_guidlines_2011, metropolis_procedural_modeling_2011, qi2018human, greg_thesis, procthor, proc4gem, wang2023gen}. Recent works incorporate priors from large language models (LLMs) or vision-language models (VLMs) \cite{robogen, katara2023gen2sim, genusd2024, pun2025hsmhierarchicalscenemotifs}. Others aim to extract 3D scenes directly from 2D images \cite{chen2024urdformer, yao2025castcomponentaligned3dscene, engstler2025syncitytrainingfreegeneration3d}, moving toward generating large-scale 3D datasets from internet-scale image corpora.
A separate line of work trains generative models that learn object relationships directly from scene data, without relying on handcrafted rules or LLMs \cite{atiss, sceneformer, legonet, diffuscene, physcene, MiDiffusion}. These models typically operate in SE(2), assuming floor-aligned layouts composed of large furniture items.
We combine the strengths of both directions by treating procedural and image-to-3D pipelines as data sources for training a generative scene model. Rather than using these pipelines at inference time, we distill their output into a flexible prior that can be adapted to downstream tasks. Our framework is agnostic to the (object ID, SE(3) pose) data source and can be augmented with real-world scenes when available.
\\
Prior generative models often represent scenes as floor-aligned SE(2) layouts and focus on static furniture arrangements \cite{atiss, sceneformer, legonet, diffuscene, MiDiffusion}. In contrast, we target cluttered SE(3) scenes composed of small, manipulable objects relevant to robotic manipulation. Many such scenes require vertical translation (e.g., placing an object on a shelf) and full 3D rotation (e.g., standing cutlery in a utensil crock), which SE(2) cannot represent. These manipulation-ready settings demand physically feasible placements, including non-penetration and static equilibrium. PhyScene \cite{physcene} encourages such feasibility through classifier-based guidance, but may still produce invalid samples. In contrast, we guarantee physical correctness via a nonlinear programming projection and simulation.
\\
In practice, the distribution of available training data often does not reflect downstream objectives, such as maximizing robot performance or aligning with human preferences. While diffusion models are typically trained to maximize likelihood under the training distribution, this is insufficient when the data does not cover task-relevant domains. We study three complementary strategies for steering a pretrained scene model toward downstream goals.
First, we adopt reinforcement learning-based post training, which has been applied in NLP and vision to optimize for user preferences \cite{gpt3_rlhf_2022, kumar2025llmposttrainingdeepdive, black2023ddpo, zhang2024largescalereinforcementlearningdiffusion}, but remains unexplored for scene generation.
Second, we explore conditional generation, widely used in SE(2) scene models, in the SE(3) setting.
Third, we introduce an inference-time Monte Carlo Tree Search (MCTS) procedure over partial scenes. Together, these tools enable goal-directed scene generation beyond the support of the original training distribution.
\\
We evaluate our generative model pipeline on five scene types ranging from tabletop to room-scale environments, compare against SE(2)-based baselines extended to SE(3), and show that the generated scenes can be used directly for robot data generation. We demonstrate post training and inference-time search using physical feasibility and high-clutter objectives relevant to robotics \cite{jia2024cluttergen}.
\\
\textbf{Summary of contributions.} Our main contribution is showing that a scene generative model trained on broad procedural data can be steered toward task-specific objectives, such as increasing clutter. Specifically: (1) we demonstrate how this can be achieved through reinforcement learning-based post training, conditional generation, and a novel MCTS-based inference-time search strategy for diffusion models; (2) we release our code, data, and model weights; and (3) we present a dataset comprising over 44 million unique SE(3) scenes spanning five distinct scene types, each featuring numerous small, movable objects. Individual scenes include up to 125 objects, supporting diverse and complex interaction scenarios relevant to robotic tasks, and providing a valuable benchmark for future work on SE(3) scene generation.

\section{SE(3) Scene Generation and Steering}
\label{sec:method}

We learn a generative model over scenes, where each object is selected from a known library and placed at a continuous SE(3) pose. Our method begins with data from a procedural generator (Section~\ref{sec:data_gen}), trains a diffusion model (Section~\ref{sec:object_set_diffusion}), and applies post processing to ensure physical feasibility (Section~\ref{sec:post_processing}). To steer the pretrained model toward downstream goals, we explore reinforcement learning post training, conditional generation, and inference-time search (Section~\ref{sec:adaptation}).

\subsection{Data Generation}
\label{sec:data_gen}

We train on procedurally generated scenes, but our method is agnostic to the scene generator and supports any source that outputs (object, pose) tuples. This includes future procedural pipelines as well as real-world scene data. While large-scale real-world SE(3) datasets remain scarce, building them from internet-scale image or video corpora is a promising direction \cite{chen2024urdformer, yao2025castcomponentaligned3dscene, engstler2025syncitytrainingfreegeneration3d}. In this work, we use a single procedural model \cite{greg_thesis} to generate training data. Distilling that data into a generative model yields a compact, unified, and differentiable scene prior that enables post training, conditional generation, and inference-time search—capabilities not easily supported by procedural systems. We provide additional data generation details in the \hyperref[appendix:datasets]{appendix}.

\subsection{SE(3) Scene Diffusion}
\label{sec:object_set_diffusion}

\textbf{Scene Representation.}
We represent a scene as an unordered object set $\mathcal{X} = \{\mathbf{o}_i \mid i \in \{1,\dots, N\}\}$, where $N$ is an upper bound on the number of objects \cite{diffuscene}. Each object $\mathbf{o}_i$ consists of an SE(3) pose, parameterized by a translation $\mathbf{p} \in \mathbb{R}^3$ and a rotation, represented as a 9D vector $\mathbf{R} \in \mathbb{R}^9$. While this rotation representation is used during training and diffusion, we project it onto SO(3) at sample time as in \cite{hitchhiker_guide_so3}. Each object also includes a one-hot vector $\mathbf{c} \in \{v \in \{0,1\}^C \mid \sum_i v_i = 1\}$, indexing a specific object asset from a fixed library $\mathcal{S}$ of $C$ assets. Following \cite{diffuscene}, we include an empty object in $\mathcal{S}$ to support variable-sized scenes. Our generative model learns distributions over such object sets $\mathcal{X}$.
\\
\textbf{Training Objective.}
We adopt the mixed discrete-continuous diffusion framework from \cite{MiDiffusion}. Specifically, we apply continuous diffusion \cite{ddpm} to $\mathbf{p}$ and $\mathbf{R}$ and discrete diffusion \cite{d3pm} to $\mathbf{c}$, conditioning each on the other during generation.
\\
\textbf{Model Architecture.}
Since we represent scenes as object sets $\mathcal{X}$, the denoising model $f$ should be object-order equivariant \cite{diffuscene}: for any permutation $\sigma(\cdot)$, it should satisfy $f(\sigma(\mathcal{X})) = \sigma(f(\mathcal{X}))$ \cite{deepsets}. Standard Transformers satisfy this property when positional encodings are omitted \cite{lee2019setTransformer}. We adopt the Flux architecture \cite{flux2024}, using its image branch without positional encodings to preserve equivariance. Flux offers efficient training and strong performance across domains such as images and music \cite{flux2024, fei2024fluxplaysmusic}. For mixed diffusion, we add input/output MLP projections following \cite{MiDiffusion}.

\subsection{Physical Feasibility Post Processing}
\label{sec:post_processing}

Even when trained on feasible data, generative models may produce SE(3) scenes that violate physical constraints, such as non-penetration or static equilibrium. These issues often arise from small numerical errors, e.g., due to mixed precision or slight misalignments.  
To enforce physical feasibility, we first resolve inter-object collisions by projecting continuous object translations to the nearest collision-free configuration while keeping orientations fixed to help preserve static equilibrium (see the \hyperref[appendix:projection]{appendix} for details).
We then simulate the scene in Drake~\cite{drake}, allowing unstable objects to settle under gravity. While projection removes penetrations, simulation corrects unstable configurations by adjusting full object poses, ensuring scenes are physically plausible and ready for downstream use.  
We apply simulation only after projection, as deep penetrations can cause large contact forces under rigid-body models, leading to explosive behavior.

\subsection{Steering Scene Generative Models Toward Downstream Objectives}
\label{sec:adaptation}

A key capability of scene generative models is their potential for steering generation toward downstream goals, even beyond the training distribution. We explore three complementary strategies: reinforcement learning-based post training (Section~\ref{sec:rl_post_training}), conditional generation (Section~\ref{sec:conditional_generation}), and inference-time search (Section~\ref{sec:mcts}).
The \hyperref[appendix:implementation_details]{appendix} contains additional details and a comparison of different steering methods.

\subsubsection{Post Training with Reinforcement Learning}
\label{sec:rl_post_training}

Distilling scene datasets into a differentiable model enables reinforcement learning (RL)–based post training. We adopt DDPO \cite{black2023ddpo} to fine-tune a continuous DDPM-based scene model \cite{ddpm} using task-specific rewards and apply the regularization from \cite{zhang2024largescalereinforcementlearningdiffusion} to stabilize training.
We use object count (clutter) as a downstream reward to test whether RL-based post training can adapt the model beyond the training distribution. To enable compatibility with existing fine-tuning methods, we use a fully continuous diffusion model, representing object categories and poses as continuous variables as in \cite{diffuscene}. Our aim is not to propose a new RL algorithm but to demonstrate the \emph{feasibility and utility} of post training for scene models.

\subsubsection{Conditional Generation}
\label{sec:conditional_generation}

Learned generative models support flexible conditioning, unlike most procedural systems \cite{procthor, greg_thesis, proc4gem}. Our models can be conditioned on language, partial scenes, or other modalities. We explore two strategies: (1) conditional training and (2) test-time inpainting using an unconditional model.
\\
\textbf{Text-conditioned generation.}
We encode prompts using BERT \cite{bert} and inject the resulting embeddings into the conditional branch of our Flux-based architecture. To enable a single model to support both conditional and unconditional generation, we randomly mask the conditioning information during training. This allows us to apply classifier-free guidance (CFG) \cite{ho2022classifierfreediffusionguidance} at inference time by interpolating between the conditional and unconditional outputs.
Our prompts are procedurally generated and can describe object counts, object identities, or spatial relationships between objects.
\\
\textbf{Scene completion and re-arrangement.}
We perform inpainting directly in the structured scene representation. Given a binary inpainting mask indicating which parts of the scene to synthesize, we generate missing content while clamping the rest to their fixed values during the reverse diffusion process \cite{sohldickstein2015deepunsupervisedlearningusing}. For example, we can rearrange scenes by regenerating the continuous poses while keeping the object categories fixed. For scene completion, we synthesize both categories and poses for empty objects. This enables consistent, plausible generation from partial inputs.

\subsubsection{Inference-Time Search via MCTS}
\label{sec:mcts}

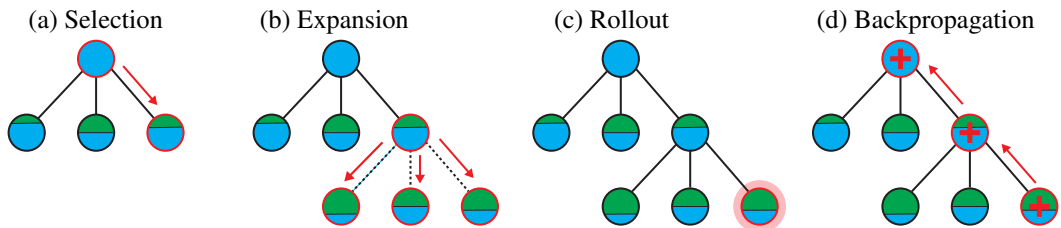
\begin{figure}
    \centering
    \begin{overpic}[width=\linewidth]{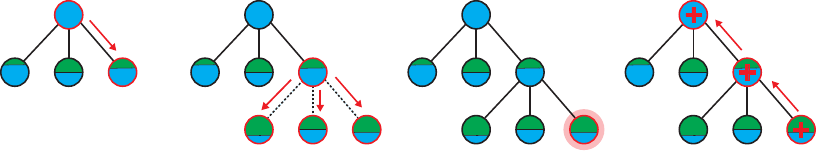}
        \put (2, 19.5) {(a) Selection}
        \put (24, 19.5) {(b) Expansion}
        \put (52, 19.5) {(c) Rollout}
        \put (77, 19.5) {(d) Backpropagation}
    \end{overpic}
    \caption{\textbf{Our MCTS inference-time search.} The root node is fully masked (blue), and child nodes represent partially inpainted scenes (blue-green). The rollout node is highlighted with a red halo.}
    \label{fig:mcts}
\end{figure}

Generative scene models can be steered toward downstream objectives at inference time. We demonstrate this via a Monte Carlo Tree Search (MCTS) procedure that incrementally constructs a scene through conditional inpainting.
At each step, an inpainting mask identifies which objects to regenerate, such as unstable ones or empty slots, and a reward function evaluates the resulting scene.
As a running example, we consider the objective of maximizing the number of physically feasible objects, i.e., objects that are non-penetrating and in static equilibrium.
Each node in the MCTS tree represents a partially completed scene and a corresponding inpainting mask.
\\
The search proceeds through the standard MCTS phases \cite{browne2012survey} (shown in Figure~\ref{fig:mcts}):
\\
\textbf{(a) Selection.}  
We traverse the tree from root to leaf, selecting at each step the child with the highest UCT \cite{uct} value: $\mathrm{UCT}(j) = \bar{r}_j + c \cdot \sqrt{\frac{2 \ln n_{\text{parent}}(j)}{n_j}}$, where $\bar{r}_j$ is the average reward of child $j$, $n_{\text{parent}}(j)$ and $n_j$ are the visit counts of parent and child, and $c$ is an exploration constant.
\\
\textbf{(b) Expansion.}  
At a leaf, we sample $B$ completions, where $B$ is the branching factor, by inpainting the masked objects with different noise initializations. Each resulting scene is evaluated to identify remaining invalid or incomplete objects, producing a new inpainting mask (e.g., flagging newly unstable or empty objects). These (scene, mask) pairs form new child nodes.
\\
\textbf{(c) Rollout.}  
One of the new children is selected randomly and scored using a task-specific reward, in our example, the number of physically feasible objects.
Since each node corresponds to a complete scene (when discarding masked objects), rollout in our setting reduces to directly reading the reward, rather than ``rolling out`` to a terminal state.
\\
\textbf{(d) Backpropagation.}  
The reward is propagated up the tree, updating average reward estimates and visit counts along the way.

We run the search for a fixed number of iterations or until a scene with no masked objects is found. If no such fully valid scene is produced, we return the best partial scene encountered, discarding any objects that remain masked.

\textbf{Controlling the Objective.}  
Our framework adapts to diverse downstream goals through two modular components: the \emph{mask generator}, which determines which objects to inpaint, and the \emph{reward function}, which evaluates scene quality. These components can be defined independently. For example, one might mask all physically invalid or empty objects, but optimize for a more targeted, yet aligned reward, such as the number of edible objects or the degree of prompt alignment. This decoupling enables flexible and modular search strategies across various downstream objectives.
\\
\textbf{Connection to Prior Work.}  
When the branching factor $B = \infty$, our method reduces to \emph{Random Search} from \citet{ma2025inferencetimescalingdiffusionmodels}, repeatedly sampling new scenes without building on previous ones.

\section{Experimental Evaluation}
\label{sec:experimental_evaluation}

\begin{figure}
    \centering
    \includegraphics[width=1\linewidth, keepaspectratio]{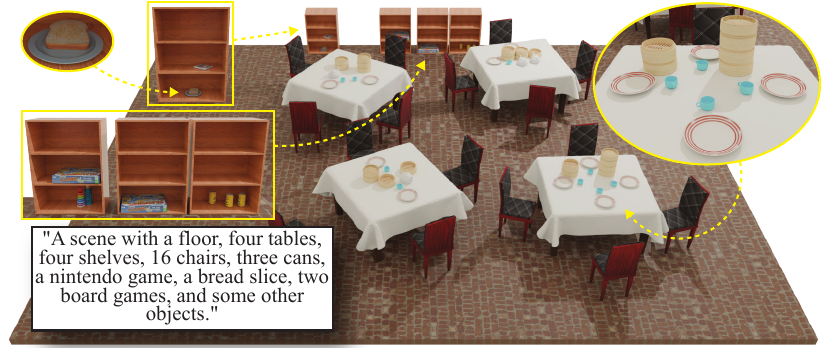}
    \caption{\textbf{Text-conditioned scene generation.} A model trained on the Restaurant (High-Clutter) dataset is queried with the shown text prompt. The generated scene matches both the large-scale layout and fine-grained object details.}
    \label{fig:text_conditioning}
\end{figure}

\subsection{Evaluation Setup}

\begin{figure}
    \centering
    \includegraphics[width=1\linewidth, keepaspectratio]{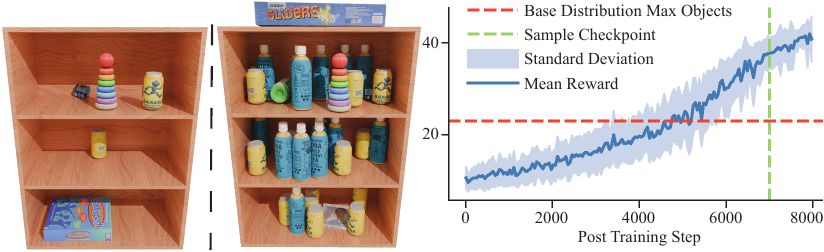}
    \caption{\textbf{RL post training with an object count reward.}
    We post-train a model originally trained on the Living Room Shelf dataset.
    \textit{Left}: Sample before post training.
    \textit{Middle}: Sample after post training.
    \textit{Right}: Reward curve. The red line marks the maximum number of objects seen during pre-training (23). Before post training, we increase the maximum number of objects allowed by the scene representation by 20 to enable higher object counts. The green line indicates the checkpoint used for sampling (step 7000), chosen to avoid overoptimization.
    }
    \label{fig:rl_results}
\end{figure}

\textbf{Metrics.}
We evaluate generative quality using image-based metrics adapted to SE(3) scenes. Following prior work on SE(2) scenes \cite{atiss, sceneformer, diffuscene, MiDiffusion}, we compute Fréchet Inception Distance (FID) and classifier accuracy (CA, in \%) based on semantic renderings. A CA near 50\% indicates realistic generation, while a CA near 100\% indicates clear separability. We render each scene from a manually defined informative viewpoint specific to the scene type.
We also report KL divergence between object category distributions, prompt-following accuracy (APF) for count and object-type prompts, and median total penetration (MTP) to assess physical feasibility. MTP is computed before applying our post processing. Full metric definitions are provided in the \hyperref[appendix:metrics]{appendix}.
\\
\textbf{Baselines.}
We compare our proposed approach against two state-of-the-art diffusion-based scene synthesis methods: (1) DiffuScene \cite{diffuscene}, which uses a 1D U-Net with a continuous DDPM objective, and (2) MiDiffusion \cite{MiDiffusion}, a Transformer-based model that employs a mixed discrete-continuous diffusion objective. We apply minimal modifications to both implementations to support our scene representation $\mathcal{X}$. For MiDiffusion, we replace floor plan conditioning with text conditioning. All models, including ours, use the same BERT text encoder \cite{bert}.
\\
\textbf{Datasets.}
As described in Section~\ref{sec:data_gen}, we generate our training data using the procedural scene generation framework from \cite{greg_thesis}. We reuse the \emph{Dimsum Table} scene type from \cite{greg_thesis} and define four additional scene types: \emph{Breakfast Table}, \emph{Living Room Shelf}, \emph{Pantry Shelf}, and \emph{Restaurant}. \emph{Restaurant} is a room-level composition that integrates \emph{Dimsum Table} and \emph{Living Room Shelf} scenes along with additional objects. For greater diversity, we split the \emph{Breakfast Table} and \emph{Restaurant} scenes into low- and high-clutter variants, reflecting the procedural generation parameters used.
In total, we sample more than 44 million SE(3) scenes across all scene types, significantly surpassing the scale of prior SE(2) scene datasets, such as 3D-FRONT \cite{fu20213d}, which contains 18,968 scenes.
The appendix provides the full set of quantitative and qualitative results across all datasets.

\subsection{Unconditional Generation}

\begin{table}[h]
\centering
\caption{Unconditional generation results on the Restaurant (High-Clutter) and Living Room Shelf datasets. * indicates that we adjusted the methods for compatibility with our scene representation.}
\label{table:unconditional-restaurant-lrshelf}
\resizebox{\textwidth}{!}{%
\begin{tabular}{lccccccccc}
\toprule
\multirow{2}{*}{Method} 
& \multicolumn{4}{c}{Restaurant (High-Clutter Variant)} 
& \multicolumn{4}{c}{Living Room Shelf} \\
\cmidrule(lr){2-5} \cmidrule(lr){6-9}
& CA (50 it, \%) $\downarrow$ & KL ($\times 10^4$) $\downarrow$ & FID $\downarrow$ & MTP (cm) $\downarrow$
& CA (100 it, \%) $\downarrow$ & KL ($\times 10^4$) $\downarrow$ & FID $\downarrow$ & MTP (cm) $\downarrow$ \\
\midrule
DiffuScene* \cite{diffuscene}     & 84.81 ± 6.49 & \textbf{0.55}  & 1.39 & 18.11 & 71.73 ± 0.99 & 4.67 & 2.18 & 0.05 \\
MiDiffusion* \cite{MiDiffusion}   & 78.63 ± 9.79 & 1.01  & 1.34 & 8.80  & 64.13 ± 1.87 & 2.51 & \textbf{2.09} & 0.03 \\
Ours                              & \textbf{70.74 ± 8.05} & 0.87  & \textbf{1.31} & \textbf{6.31} & \textbf{52.84 ± 1.26} & \textbf{2.13} & \textbf{2.09} & \textbf{0.02} \\
\bottomrule
\end{tabular}
}
\end{table}

We report unconditional generation results for the Restaurant (High-Clutter) and Living Room Shelf datasets in Table~\ref{table:unconditional-restaurant-lrshelf}; additional results, including samples from a single model jointly trained across all datasets, are provided in the \hyperref[appendix:additional_results]{appendix}.
Rather than training separate unconditional models, we use our text-conditioned models by providing empty conditioning inputs at sampling time. 
Our model achieves strong FID and significantly lower MTP compared to baselines, indicating that it produces scenes that are both visually realistic and physically plausible. Classifier accuracy (CA) closer to 50\% further supports that our samples are harder to distinguish from dataset scenes.
While we do not always achieve the lowest KL divergence, all methods obtain very low KL values on our datasets. Since KL is near saturation, the differences are minor, and this metric is less informative in our setting; therefore, we report it primarily for completeness, following prior work.

\subsection{Post Training with Reinforcement Learning}

We apply reinforcement learning (RL) post training to a model trained on the Living Room Shelf dataset, using an object count reward. Figure~\ref{fig:rl_results} shows the reward curve and sample scenes before and after post training. We choose a checkpoint before overoptimization occurs to maintain scene quality. Additional results are provided in the \hyperref[appendix:additional_results]{appendix}.
RL-based post training successfully adapts the pretrained model to generate scenes with object counts substantially exceeding those observed during pretraining. By expanding the maximum object capacity in the scene representation before post training, we enable the model to extrapolate beyond its original range without requiring retraining from scratch. This demonstrates that post training can effectively shift and reshape scene distributions toward task-specific goals.

\subsection{Conditional Generation}

\begin{table}[h]
\centering
\caption{Conditional generation results on the Breakfast Table (High-Clutter) and Pantry Shelf datasets. * indicates that we adjusted the methods for compatibility with our scene representation.}
\label{table:conditional-breakfast-pantry}
\resizebox{\textwidth}{!}{%
\begin{tabular}{lccccccccc}
\toprule
\multirow{2}{*}{Method} 
& \multicolumn{4}{c}{Breakfast Table (High-Clutter Variant)} 
& \multicolumn{4}{c}{Pantry Shelf} \\
\cmidrule(lr){2-5} \cmidrule(lr){6-9}
& CA (50 it, \%) $\downarrow$ & KL ($\times 10^4$) $\downarrow$ & FID $\downarrow$ & APF $\uparrow$
& CA (50 it, \%) $\downarrow$ & KL ($\times 10^4$) $\downarrow$ & FID $\downarrow$ & APF $\uparrow$ \\
\midrule
DiffuScene* \cite{diffuscene}     & 82.38 ± 3.82 & 0.96 & 1.87 & 0.76 & 84.65 ± 2.23 & 0.91 & 1.93 & 0.88 \\
MiDiffusion* \cite{MiDiffusion}   & 82.22 ± 3.11 & 0.58 & 1.93 & 0.60 & 87.24 ± 1.80 & \textbf{0.64} & 1.89 & 0.72 \\
Ours                              & \textbf{68.44 ± 4.67} & \textbf{0.30} & \textbf{1.84} & \textbf{0.86} & \textbf{82.78 ± 3.44} & 0.66 & \textbf{1.88} & \textbf{0.98} \\
\bottomrule
\end{tabular}
}
\end{table}

\begin{figure}
    \centering
    \includegraphics[width=1\linewidth, keepaspectratio]{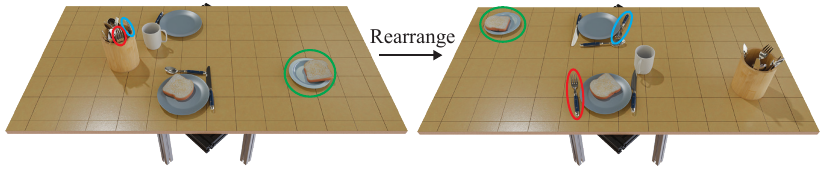}
    \caption{\textbf{Scene rearrangement example.} A scene from the Restaurant (Low-Clutter) dataset is rearranged via inpainting by a model trained on the same dataset. Red, green, and blue ellipses highlight corresponding objects. Notably, cutlery is moved from the utensil crock to the table, requiring full SO(3) rotation modeling.}
    \label{fig:scene_completion_and_rearrangement}
\end{figure}

\begin{figure}
    \centering
    \includegraphics[width=1\linewidth, keepaspectratio]{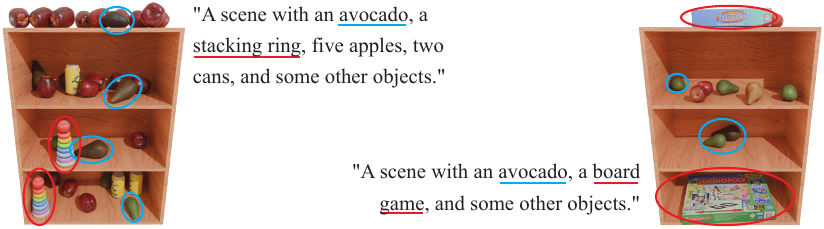}
    \caption{\textbf{Interpolation between Living Room and Pantry Shelf Scenes.} We train a joint model on both datasets with a 50/50 batch mix. By prompting for objects unique to each dataset (red = Living Room Shelf, blue = Pantry Shelf), we guide the model to generate interpolated scenes.}
    \label{fig:shelf_interpolation}
\end{figure}

We report quantitative results for text-conditioned generation on the Breakfast Table (High-Clutter) and Pantry Shelf datasets in Table~\ref{table:conditional-breakfast-pantry}.
Figure~\ref{fig:text_conditioning} shows examples of text-conditioned generation, and Figure~\ref{fig:scene_completion_and_rearrangement} illustrates scene rearrangement via partial inpainting.
Additional results, including scene completion, are provided in the \hyperref[appendix:additional_results]{appendix}.  
Our model outperforms baselines in CA, FID, and APF, indicating stronger prompt adherence and overall generation quality. Qualitative examples further show that our model captures both large-scale layouts and fine-grained object details.
\\
\textbf{Cotraining across scene types.}
We also investigate whether cotraining on the Living Room Shelf and Pantry Shelf datasets enables interpolation.  
During training, we use equal batch mixing ratios across sub-datasets. As shown in Figure~\ref{fig:shelf_interpolation}, prompting the model with mixed object descriptions from both environments leads to interpolated scenes that combine elements of each dataset, demonstrating that the model captures a meaningful joint distribution.

\subsection{Inference-Time Search}

\begin{figure}
    \centering
    \includegraphics[width=1\linewidth, keepaspectratio]{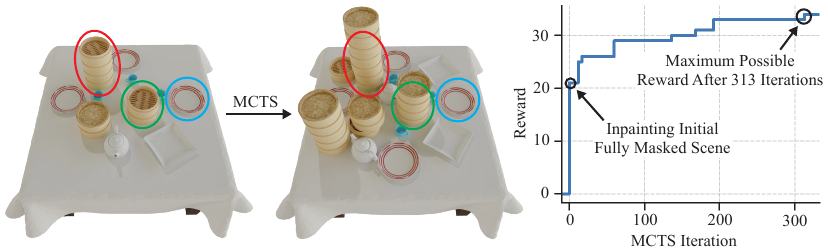}
    \caption{\textbf{Inference-time MCTS.}
    We apply MCTS at inference time to generate a Dimsum scene that maximizes the number of physically feasible objects.
    \textit{Left}: Initial sample and final result after search. Red, green, and blue ellipses highlight corresponding objects. Note how the search completes the steamer stacks.
    \textit{Right}: Reward curve. Inpainting the fully masked scene (equivalent to unconditional sampling) yields 21 feasible objects in the best of $B=3$ samples. MCTS reaches the maximum possible 34 objects after 313 iterations, with reward rising quickly, then plateauing.
    }
    \label{fig:search_results}
\end{figure}

Figure~\ref{fig:search_results} shows how MCTS optimizes the number of physically feasible objects in a Dimsum scene with a branching factor $B=3$. The training set had a mean of 17.1 objects and a maximum of 34; MCTS reaches this maximum, demonstrating that inference-time search can push scene complexity well beyond typical training-time levels. 
Scene distributions encode local structure, for example, steamers often appear in vertical stacks. Our results show that the model captures such patterns: MCTS incrementally builds realistic, physically feasible stacks by exploiting inductive biases learned during pretraining, without requiring retraining.

\section{Conclusion}
\label{sec:conclusion}

We presented a diffusion-based framework for SE(3) scene generation that distills large-scale procedural data into a flexible, physically grounded prior. Our model predicts object categories from a fixed asset library and continuous poses, and supports adaptation via RL-based post training, conditional generation, and inference-time search.  
Experiments across five scene types show that the pretrained model enables strong unconditional and conditional generation, that post training improves targeted metrics such as clutter, and that MCTS search can optimize task rewards without retraining. To qualitatively validate simulation readiness, we imported generated scenes into the Drake simulator and successfully teleoperated a mobile KUKA iiwa robot to perform pick-and-place interactions without requiring manual scene corrections (see Figure~\ref{fig:teaser} and \href{https://steerable-scene-generation.github.io/}{supplementary videos}).  
Together, these results demonstrate that a single model can flexibly adapt scene distributions without handcrafted tuning or retraining.  
Our work highlights a scalable approach to robotic scene generation: pretraining on broad data sources, then steering toward task-specific goals.

\section{Limitations}
\label{sec:limitations}

While our method demonstrates the feasibility and benefits of steering scene generative models toward downstream objectives, several limitations remain. First, although procedural data provides scalable supervision, it may not fully capture the complexity and variability of real-world environments. Incorporating real-world SE(3) datasets, potentially extracted from internet-scale image or video corpora, remains an important direction for enhancing realism.
Second, we adopt fully continuous diffusion models to enable reinforcement learning-based post training, rather than our full mixed discrete-continuous models. We leave applying post training to mixed discrete-continuous settings as future work. Additionally, we observe that when post training pushes object count close to the maximum allowed by the scene representation, overoptimization can occur: samples exhibit many objects but no longer resemble the original data distribution. While expanding the maximum object capacity helps, fully continuous models still struggle to maintain quality when handling many additional object tokens, limiting the effectiveness of this strategy.
Third, our object representation uses a fixed asset library with one-hot encodings, reflecting a practical design choice aligned with robotics workflows, which often depend on pre-validated simulation assets to ensure high-quality geometry and physical properties for realistic simulations. While this limits generalization to novel geometries without retraining, it enables precise control over the asset set, and our steering methods remain compatible with alternative object representations.
Fourth, while our object library currently consists of rigid bodies, it naturally extends to articulated objects (e.g., drawers, cabinets) without requiring changes to the method. We leave the exploration of articulated scenes for future work.
Fifth, our adaptation strategies—post training, conditional generation, and inference-time search—are proof-of-concept demonstrations. Future work could explore more sophisticated reward functions, conditioning schemes, and search objectives tailored to specific robot tasks.
Finally, while we demonstrate simulation-readiness via teleoperation, scaling to large-scale autonomous robot training across generated scenes is an important direction for future work.

\acknowledgments{This work was partially supported by Amazon.com, PO No. 2D-15693043. The Toyota Research Institute partially supported this work. This article solely reflects the opinions and conclusions of its authors and not TRI or any other Toyota entity.
We thank Rick Cory, Ge Yang, Phillip Isola, Adam Wei, and Jeremy Binagia for fruitful discussions.

}

\bibliography{ref}  %

\appendix

\section{Procedurally Generated Datasets}
\label{appendix:datasets}

\begin{table}[h]
\centering
\caption{Dataset statistics for each scene type: number of unique object assets, minimum, maximum, and mean number of objects per scene, and total number of scenes. The final row reports combined statistics across all datasets and corresponds to the dataset used for cotraining across all scene types.}
\label{table:dataset_stats}
\begin{tabular}{lrrrrr}
\toprule
\multirow{2}{*}{Dataset} & \# Object & Min & Max & Mean & Total \\
                         & Assets    & Objects & Objects & Objects & Scenes \\
\midrule
Breakfast Table (Low-Clutter) & 14 & 4 & 26 & 8.87 & 6,679,981 \\
Breakfast Table (High-Clutter) & 15 & 4 & 46 & 20.96 & 2,655,407 \\
Dimsum Table & 7 & 7 & 34 & 18.14 & 13,042,071 \\
Living Room Shelf & 18 & 4 & 23 & 5.93 & 9,087,513 \\
Pantry Shelf & 14 & 3 & 61 & 27.98 & 6,171,940 \\
Restaurant (High-Clutter) & 27 & 33 & 125 & 60.54 & 1,195,751 \\
Restaurant (Low-Clutter) & 27 & 9 & 107 & 35.08 & 1,797,170 \\
\midrule
\textit{All Datasets Combined} & 46 & 3 & 125 & 17.17 & 44,790,942 \\
\bottomrule
\end{tabular}
\end{table}

\begin{figure}
\centering
\includegraphics[width=\linewidth, keepaspectratio]{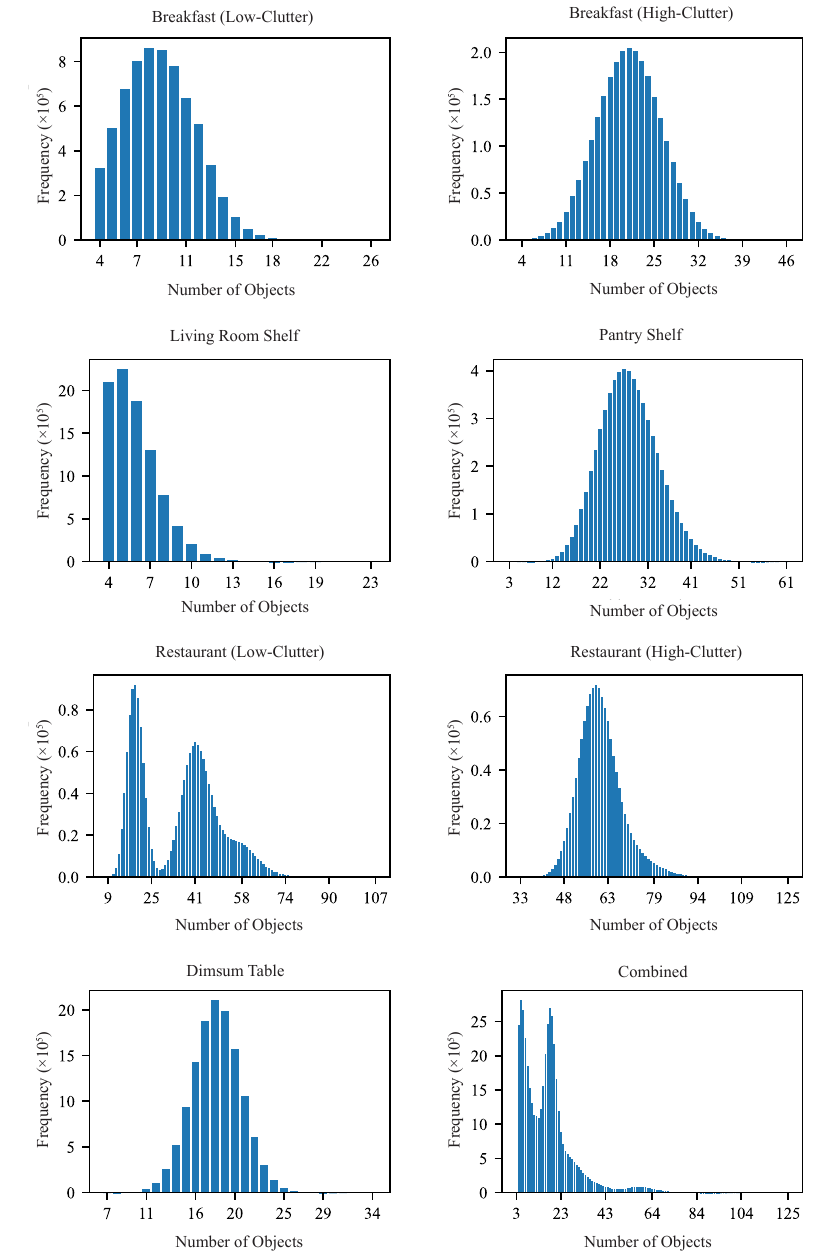}
\caption{\textbf{Dataset statistics.} The object count statistics for each dataset and the combined dataset. The first x-label on each histogram represents the minimum object number, and the last one represents the maximum object number. Note the long tails of the distributions.}
\label{fig:datasets_stats}
\end{figure}

We generate a large-scale SE(3) scene dataset to address the lack of existing resources with full 6-DOF object annotations. Prior scene generative models primarily relied on the 3D-FRONT dataset~\cite{fu20213d}, which provides SE(2) room layouts for 18,968 scenes and includes 13,151 unique furniture object assets. Most prior work~\cite{atiss, diffuscene, MiDiffusion} focused on the bedroom, dining, and living room subsets, where scenes contain between 3 and 21 objects. After preprocessing, these subsets typically yield 4,041 bedroom, 900 dining room, and 813 living room scenes.

In contrast, our dataset includes full SE(3) object poses and dramatically increases the number and variety of sampled arrangements. We generate a total of 44,790,942 scenes across five scene types, with the largest (\emph{Dimsum Table}) containing over 13 million scenes. While our dataset uses only 46 unique object assets (fewer than 3D-FRONT), it achieves substantial variation through randomized object counts and physically diverse placements.

Table~\ref{table:dataset_stats} summarizes key statistics, and Figure~\ref{fig:datasets_stats} shows object count histograms for each dataset.

\subsection{Procedural Data Generation}

We adopt the probabilistic context-free scene grammars from \citet{greg_thesis} to generate our datasets. These grammars define procedural recipes for constructing scenes as sequences of object types and production rules, and implicitly encode a tree structure in which nodes correspond to objects and edges represent recursive expansion rules.

Sampling begins from a root node (e.g., a table), which stochastically spawns child nodes based on the grammar. For example, a table might generate between one and $P$ place settings, where $P$ is drawn from a discrete distribution and each object pose is sampled from a continuous distribution (e.g., uniform or Gaussian). These place settings can recursively spawn sub-objects such as plates or cutlery, which may in turn generate additional items (e.g., food) until terminal nodes are reached.

As context-free grammars cannot enforce cross-branch constraints (e.g., non-penetration between unrelated objects), \citet{greg_thesis} introduced constraints such as physical feasibility via rejection sampling and Hamiltonian Monte Carlo.

We implement one grammar per scene type using the original codebase. However, sampling can be slow for complex scenes, particularly high-clutter environments like our \emph{Restaurant (High-Clutter)} dataset, where many proposed scenes are rejected due to constraint violations. To scale up generation, we parallelize sampling across 192 CPUs on AWS EC2 M7A Metal 48xlarge instances. In the worst case, we achieved a throughput of approximately 60,000 valid scenes per day per instance. More efficient sampling may be possible with improved grammar design or a dedicated distributed sampling system.
Once trained, our generative model supports scene sampling at significantly higher throughput than the procedural generation process.

\subsection{Text Prompt Generation}

We generate textual annotations for our procedurally created scenes using a rule-based system, similar to the one used in \cite{diffuscene}, that maps scene content into natural language descriptions. Each scene is annotated with one or more sentences drawn from three annotation types: object count, object names, and spatial relationships.

\textbf{Object Count.}
This annotation describes the total number of non-empty objects in the scene, e.g., ``A scene with 32 objects.``

\textbf{Object Names.}
We extract object names from the scene and generate a descriptive sentence listing either all objects or a sampled subset, e.g., ``A scene with a plate, two bowls, and some other objects.`` In room-scale scenes, we prioritize large objects such as tables or shelves when sampling subsets. We apply heuristics for article selection and pluralization (e.g., ``a bowl`` vs. ``an apple``), and express counts using word forms for numbers up to ten (e.g., ``two bowls``) and numerals for larger quantities (e.g., ``12 bowls``).

\textbf{Spatial Relationships.}
We extend the object name sentence with spatial relationships derived from the 3D positions of mentioned objects. These include topological (e.g., ``on top of``), directional (e.g., ``to the left of``), and containment-based (e.g., ``inside``) relations. Only pairs within a distance threshold are considered, and relationships are expressed only between named objects. For shelf and room scenes, we apply scene-specific rules, such as identifying shelf tiers or prioritizing furniture. To disambiguate repeated object types, we insert ordinal labels where necessary (e.g., ``the second apple``).

To encourage diversity in annotations, we duplicate each dataset scene three times and generate a different annotation type for each copy.

\section{Implementation Details}
\label{appendix:implementation_details}

In this section, we go over some of the key implementation details.
The complete codebase is available at \url{https://github.com/nepfaff/steerable-scene-generation}.

\subsection{Model Architecture}

\begin{figure}
\centering
\includegraphics[width=\linewidth, keepaspectratio]{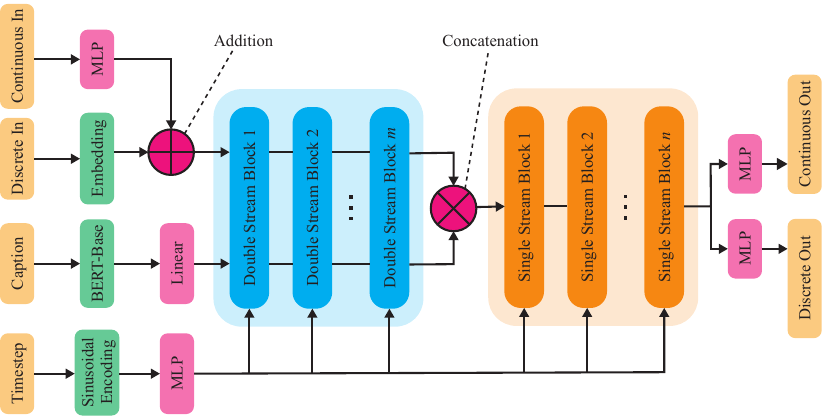}
\caption{\textbf{Denoising network architecture.} Our model follows a Flux-style design~\cite{flux2024}, operating over mixed discrete and continuous scene representations. Discrete and continuous inputs are separately embedded and added, then passed through double stream transformer blocks that separately process scene and text features before merging at the attention step. After concatenation along the token dimension, single stream transformer blocks further process the fused representation. Final outputs are decoded into discrete and continuous components for the next timestep.}
\label{fig:denoising_network}
\end{figure}

We adopt a Flux-style architecture~\cite{flux2024} using its image branch without positional encoding to preserve object order equivariance (see Figure~\ref{fig:denoising_network}). 
Our denoising network takes as input the discrete component $c_t$ and continuous components $\mathbf{R}_t$ and $\mathbf{p}_t$ at timestep $t$, and predicts the corresponding outputs $c_{t-1}$, $\mathbf{R}_{t-1}$, and $\mathbf{p}_{t-1}$ for the next step.

The discrete and continuous inputs are embedded separately and added together following \citet{MiDiffusion}.  
The result is fed into double stream transformer blocks that independently process scene and text features using separate weights, merging during the attention operation.  
After processing by double stream blocks, the modalities are concatenated along the token dimension and passed through single stream transformer blocks.  
The final outputs are projected via separate multilayer perceptrons (MLPs) into discrete and continuous components.

We use a frozen BERT-Base encoder~\cite{bert} for text features, and sinusoidal embeddings for the timestep, which are injected via modulation layers.

\subsection{Model Training}

We train all models using exponential moving average (EMA) updates~\cite{tarvainen2018meanteachersbetterrole} and an AdamW optimizer~\cite{loshchilov2019decoupledweightdecayregularization} with a learning rate of 2.0e-4 and weight decay of 1.0e-3. We apply a cosine learning rate schedule with linear warmup over the first 5000 iterations. Training uses a linear diffusion noise schedule with 1000 timesteps for both training and inference. To enable classifier-free guidance (CFG), we drop the conditional input with 10\% probability during training.

Our denoising network utilizes 5 double-stream and 10 single-stream Transformer blocks, resulting in a total of 88.3 million parameters. Text prompts are encoded using 110 BERT~\cite{bert} tokens, corresponding to the longest caption across our datasets. All models are trained for 200 thousand iterations with a batch size of 256 per GPU.

Training is performed on 8 NVIDIA A100 GPUs. Most experiments use 40GB A100s, while the Restaurant and cotraining experiments use 80GB A100s. We use full precision during final training to accurately distinguish large translation values in scene poses, which low-precision formats cannot reliably separate. bfloat16 was used during development for faster experimentation and works well for non-room-level datasets with small translation ranges. We provide a training precision ablation in Section~\ref{appendix:ablations}.  
Overall training times range from approximately 1.06 days for the Dimsum Table dataset to 1.65 days for cotraining across all datasets.

\subsection{Cotraining Recipe}
\label{appendix:cotraining_recipe}

During cotraining, we sample batches such that each scene type occurs with equal probability, ensuring balanced exposure to all subdatasets. We maintain a separate infinite iterable stream for each subdataset, using random offsets and buffered shuffling to enhance diversity during sampling. At each step, a subdataset is selected uniformly at random, and a scene is drawn from its corresponding iterator.

We apply this cotraining setup both for shelf cotraining, where the model is jointly trained on the Living Room Shelf and Pantry Shelf datasets, and for full cotraining across all five scene types.

\subsection{Physical Feasibility Post Processing}
\label{appendix:projection}

To enforce non-penetration, we apply a projection step that adjusts object translations while keeping orientations fixed. We solve the following nonlinear optimization problem:
\begin{equation}
    \min_{\mathbf{p}} \|\mathbf{p} - \mathbf{p}_0\|_2^2 \quad
    \text{s.t.} \quad d(i,j) \ge 0, \quad \forall i,j \in \{1,\dots,N\},
\end{equation}
where $\mathbf{p}$ and $\mathbf{p}_0$ denote the optimized and original object translations, respectively, and $d(i,j)$ is the signed distance between objects $i$ and $j$ as computed by a collision checker.

We solve this optimization using the SNOPT solver~\cite{snopt}. 
In some complex scenes, particularly in the Restaurant dataset, SNOPT may fail to converge due to the presence of many complicated nonlinear non-convex constraints, where the objects are in contact without penetration. Finding solutions satisfying such constraints can be challenging for general-purpose nonlinear solvers, especially when using tight numerical tolerances.

To assess the effectiveness of our post processing, we conduct a qualitative ablation comparing scenes generated with (1) no post processing, (2) simulation-only, and (3) projection followed by simulation (see Section~\ref{appendix:ablations}). We find that combining projection and simulation yields the most physically plausible scenes, especially in cluttered environments.

\subsection{Post Training with Reinforcement Learning}
\label{appendix:rl_post_training}

We fine-tune our continuous diffusion model using DDPO~\cite{black2023ddpo}, treating the denoising process as a multi-step decision-making problem. We adopt the score function variant (DDPO$_{\text{SF}}$), which uses the REINFORCE estimator without a learned value function or importance sampling.

\textbf{Trajectory generation.}
We use a DDIM \cite{ddim} scheduler with 100–150 steps to generate full denoising trajectories. Gradients are computed either across all timesteps or a uniformly sampled subset, following the strategy in~\cite{zhang2024largescalereinforcementlearningdiffusion}. To maintain consistency between training and inference, we use the same scheduler and number of steps at both stages. Using mismatched settings (e.g., 200 DDIM or 1000 DDPM steps for inference after post training) still improves over the pretrained model but underperforms compared to the configuration used during post training.

\textbf{Rewards.}
We use task-specific rewards evaluated only at the final denoised scene ($t=0$). All results in this paper use object count as the reward signal. Advantages are computed by normalizing the per-batch rewards using the batch mean and standard deviation. To ensure stability in distributed settings, we synchronize these statistics across all workers before computing advantages.

\textbf{Loss.}
As a proof-of-concept, we use the standard REINFORCE objective to compute the RL loss:
\begin{equation}
    \mathcal{L}_{\text{RL}} = -\mathbb{E} \left[\sum_t \log p(x_t \mid x_{t+1}) \cdot A \right],
\end{equation}
where $A$ denotes the advantage. We apply an additional DDPM \cite{ddpm} loss term, weighted by a tunable coefficient $\lambda_{\text{DDPM}}$, to stabilize optimization~\cite{zhang2024largescalereinforcementlearningdiffusion}. We find values of $\lambda_{\text{DDPM}} \in [100, 200]$ to be effective.

Training is conducted with a batch size of 32 per GPU across 8 NVIDIA A100 80GB GPUs.

\textbf{Extension to Mixed Discrete-Continuous Models.}
The DDPO framework can be extended to the discrete D3PM setting~\cite{d3pm}, and thus also to mixed discrete-continuous diffusion models. In this case, one would need to compute gradients of the form
$\frac{\partial \log p_\theta(x_t \mid x_{t+1})}{\partial \theta}$,
which are defined for both DDPM and D3PM formulations. The resulting RL objective would then be the sum of REINFORCE losses over both the continuous (DDPM) and discrete (D3PM) components. A key challenge in this direction is developing few-step samplers for D3PMs to make optimization practical.

\subsection{Conditional Generation}
\label{appendix:conditional_generation}

\subsubsection{Text-Conditioned Generation}

We support text-conditioned generation by encoding prompts using a frozen BERT-Base encoder~\cite{bert}. The resulting embeddings are injected into the conditional input branch of our Flux-style architecture. We use 110 tokens per prompt, corresponding to the maximum prompt length across all scene types in our datasets.

To enable both conditional and unconditional generation within a single model, we randomly mask the conditioning input with 10\% probability during training. This allows classifier-free guidance (CFG)~\cite{ho2022classifierfreediffusionguidance} at inference time, where predictions are computed using a weighted combination of conditional and unconditional outputs:
\begin{equation}
    \hat{x} = (1 + w) \cdot \hat{x}_{\text{cond}} - w \cdot \hat{x}_{\text{uncond}},
\end{equation}
where $w$ is the guidance weight, and $\hat{x}_{\text{cond}}$ and $\hat{x}_{\text{uncond}}$ are the model predictions under conditional and unconditional contexts, respectively. A weight of $w = -1$ corresponds to unconditional sampling, $w = 0$ yields conditional sampling without guidance, and $w > 0$ applies classifier-free guidance during sampling. We apply CFG to both discrete and continuous components.

\subsubsection{Inpainting}

We perform structured inpainting by masking a subset of objects or object attributes and generating the missing content while preserving the unmasked parts of the scene. Inpainting operates natively over our mixed discrete-continuous representation: masked components are initialized with noise and updated via reverse diffusion, while unmasked components are clamped to their original values throughout sampling.

This approach enables flexible editing operations, such as object rearrangement (resampling poses while keeping categories fixed) and scene completion (synthesizing both categories and poses for empty object slots).

\subsection{Inference-Time Search via MCTS}
\label{appendix:inference_time_search}

Our inference-time search framework incrementally inpaints masked regions of a scene, guided by a task-specific reward. In this work, we use the number of physically feasible objects as a proof of concept. However, the framework is modular and supports alternative objectives by substituting the mask generator and reward function components.

\subsubsection{Mask Generator}
Objects are included in the inpainting mask if they are deemed \emph{invalid}, meaning they violate either (1) non-penetration or (2) static equilibrium constraints.

Penetration is detected using Drake’s \cite{drake} signed distance checker, and all objects involved in penetrating pairs are masked. Static equilibrium is evaluated by simulating the scene for 0.1s in Drake. An object is deemed unstable if its translation or rotation exceeds predefined thresholds. To avoid instability and spurious support relationships, penetrating objects are excluded from the simulation.

Some objects are predefined as \emph{welded}, such as tables in tabletop scenes and shelves in shelf scenes. These objects serve as supporting surfaces in the absence of a floor and are excluded from masking. Masking welded objects would cause all other objects to fall during simulation, severely degrading search efficiency.

\subsubsection{Reward Function}
The reward is defined as the number of \emph{physically feasible} objects in the scene—those that are both non-penetrating and statically stable according to the same criteria used in the mask generator. Since each object's feasibility is evaluated as a binary outcome, the total reward is an integer count of feasible objects.

\subsubsection{Relation to Gradient-Based Guidance}
In addition to MCTS, diffusion models can also be steered at inference time using gradient-based guidance, which relies on differentiable reward functions and typically incurs lower compute cost. In contrast, MCTS supports arbitrary non-differentiable objectives, making it more broadly applicable within our framework.

\subsection{Comparison of Steering Methods}
Our framework supports multiple complementary steering strategies:
\begin{itemize}
    \item \textbf{RL post-training} optimizes arbitrary objectives with fast inference but requires additional training.  
    \item \textbf{Conditional generation} requires conditioning labels during training and introduces modest extra training cost (e.g., encoding text), while incurring negligible overhead at inference.
    \item \textbf{Inpainting} requires no retraining but is limited to the training distribution; it is useful for rearrangement or adding distractors, and also serves as a building block for MCTS.  
    \item \textbf{MCTS} enables flexible optimization with arbitrary objectives without retraining, at the cost of slower inference.  
\end{itemize}

These strategies can be composed, for example, a conditional model can be refined with RL and further steered with MCTS, similar in spirit to modern LLM pipelines.

\section{Metric Definitions and Evaluation Details}
\label{appendix:metrics}

\begin{figure}
\centering
\includegraphics[width=\linewidth, keepaspectratio]{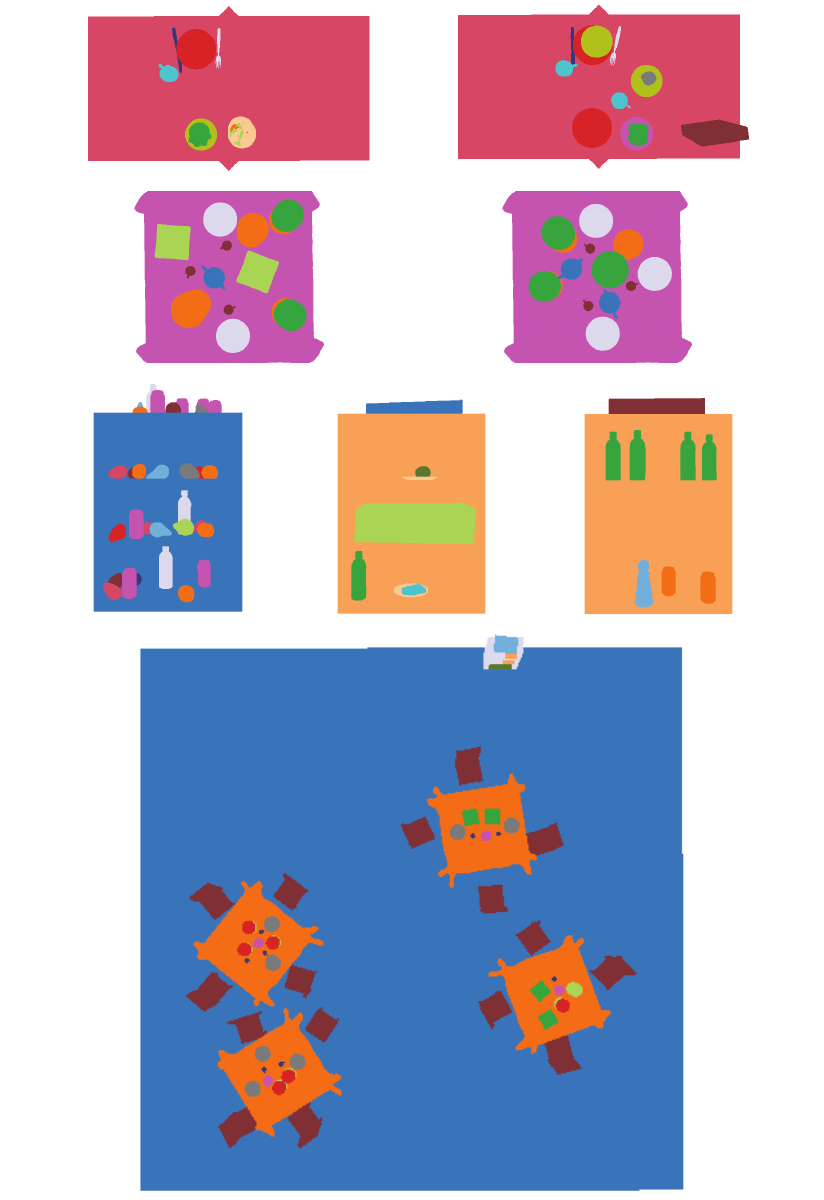}
\caption{Semantic renderings used to compute image-based metrics such as classifier accuracy (CA) and Fréchet Inception Distance (FID). Rows correspond to the Breakfast Table, Dimsum Table, Pantry and Living Room Shelves, and Restaurant datasets, respectively.}
\label{fig:semantic_renders}
\end{figure}

Prior work on floor-aligned scenes with SE(2) poses typically evaluates generative models using top-down renderings and image-based metrics such as Fréchet Inception Distance (FID) and classifier accuracy (CA, in \%) \cite{atiss, sceneformer, diffuscene, MiDiffusion}. However, these approaches are suboptimal for SE(3) scenes, where a single view may not capture the full scene state due to occlusions. For example, in a cluttered bin, no camera can reveal objects underneath the top layer.

Despite these challenges, image-based metrics remain practical proxies. Due to the lack of pretrained feature extractors for full SE(3) representations, we reuse SE(2)-based metrics, which provide sufficient sensitivity to distinguish between generative methods. Accordingly, we adopt FID and CA as primary metrics. CA is computed by training a binary classifier to distinguish between real and generated semantic renderings. A CA of 50\% indicates indistinguishable distributions, while a CA near 100\% indicates clear separability. We average CA over 10 runs with random train/test splits to reduce variance. To ensure metric sensitivity across datasets, we tune classifier strength (i.e., training iterations) per dataset; this tuning is fixed across methods for fair comparison. Scenes are rendered from a manually selected informative viewpoint at 640×480 resolution using semantic colors assigned by object class as in \cite{diffuscene}. Figure~\ref{fig:semantic_renders} shows example semantic renderings used for CA and FID evaluation.

In addition to FID and CA, we report three further metrics.  
We compute KL divergence between object category distributions following prior work \cite{atiss}; although all models achieve low KL values on our datasets, we include it for completeness.  
For conditional generation, we report average prompt-following accuracy (APF), computed for two automatically verifiable prompt types: (1) number of objects, and (2) inclusion of specified object categories.  
For unconditional generation, we report the median total penetration (MTP) measured in centimeters before applying physical feasibility post processing; after projection, MTP is guaranteed to be zero.  
All metrics are computed using 5{,}000 training scenes and 5{,}000 generated scenes.  
For conditional generation, prompts are extracted from the training scenes and reused during synthesis.

\section{Additional Results}
\label{appendix:additional_results}

\subsection{Unconditional Generation}

\begin{table}[h]
\centering
\caption{Unconditional generation results on the Pantry Shelf and Breakfast Table (Low-Clutter) datasets. * indicates that we adjusted the methods for compatibility with our scene representation.}
\label{table:appendix-unconditional-pantry-breakfast}
\resizebox{\textwidth}{!}{%
\begin{tabular}{lccccccccc}
\toprule
\multirow{2}{*}{Method} 
& \multicolumn{4}{c}{Pantry Shelf} 
& \multicolumn{4}{c}{Breakfast Table (Low-Clutter Variant)} \\
\cmidrule(lr){2-5} \cmidrule(lr){6-9}
& CA (50 it, \%) $\downarrow$ & KL ($\times 10^4$) $\downarrow$ & FID $\downarrow$ & MTP (cm) $\downarrow$
& CA (25 it, \%) $\downarrow$ & KL ($\times 10^4$) $\downarrow$ & FID $\downarrow$ & MTP (cm) $\downarrow$ \\
\midrule
DiffuScene* \cite{diffuscene}     & 84.38 ± 1.79 & 1.99 & 1.97 & 9.32 & 68.32 ± 8.48 & 2.64 & 2.59 & 5.94 \\
MiDiffusion* \cite{MiDiffusion}   & 85.42 ± 0.98 & \textbf{1.18} & 1.90 & 5.57 & 65.34 ± 7.39 & 1.79 & 2.43 & 3.81 \\
Ours                              & \textbf{84.04 ± 0.58} & 1.45 & \textbf{1.89} & \textbf{4.92} & \textbf{57.85 ± 4.98} & \textbf{1.67} & \textbf{2.39} & \textbf{3.50} \\
\bottomrule
\end{tabular}
}
\end{table}

\begin{table}[h]
\centering
\caption{Unconditional generation results on the Dimsum Table and Breakfast Table (High-Clutter) datasets. * indicates that we adjusted the methods for compatibility with our scene representation.}
\label{table:table:appendix-unconditional-dimsum-breakfastHigh}
\resizebox{\textwidth}{!}{%
\begin{tabular}{lccccccccc}
\toprule
\multirow{2}{*}{Method} 
& \multicolumn{4}{c}{Dimsum Table} 
& \multicolumn{4}{c}{Breakfast Table (High-Clutter Variant)} \\
\cmidrule(lr){2-5} \cmidrule(lr){6-9}
& CA (100 it, \%) $\downarrow$ & KL ($\times 10^4$) $\downarrow$ & FID $\downarrow$ & MTP (cm) $\downarrow$
& CA (50 it, \%) $\downarrow$ & KL ($\times 10^4$) $\downarrow$ & FID $\downarrow$ & MTP (cm) $\downarrow$ \\
\midrule
DiffuScene* \cite{diffuscene}     & 83.56 ± 8.14 & \textbf{0.39}  & 0.95 & \textbf{0.18} & 81.71 ± 5.00 & 1.57  & \textbf{1.86} & \textbf{6.97} \\
MiDiffusion* \cite{MiDiffusion}   & 78.69 ± 10.25 & 0.57  & 0.96 & 0.19 & 78.86 ± 4.89 & 0.84 & 1.943 & 7.31 \\
Ours                              & \textbf{57.66 ± 4.70} & 0.63  & \textbf{0.89} & 0.20 & \textbf{69.82 ± 5.37} & \textbf{0.81} & 1.88 & \textbf{6.97} \\
\bottomrule
\end{tabular}
}
\end{table}

\begin{table}[h]
\centering
\caption{Unconditional generation results on the Restaurant (Low-Clutter) dataset. * indicates that we adjusted the methods for compatibility with our scene representation.}
\label{table:appendix-unconditional-restaurant-low}
\resizebox{0.6\textwidth}{!}{%
\begin{tabular}{lcccc}
\toprule
\multirow{2}{*}{Method} 
& \multicolumn{4}{c}{Restaurant (Low-Clutter Variant)} \\
\cmidrule(lr){2-5}
& CA (50 it, \%) $\downarrow$ & KL ($\times 10^4$) $\downarrow$ & FID $\downarrow$ & MTP (cm) $\downarrow$ \\
\midrule
DiffuScene* \cite{diffuscene}     & 78.41 ± 8.14 & 1.60 & 1.52 & 7.22 \\
MiDiffusion* \cite{MiDiffusion}   & 75.34 ± 10.48 & \textbf{1.01} & 1.45 & 3.50 \\
Ours                              & \textbf{66.11 ± 6.93} & 1.18 & \textbf{1.35} & \textbf{2.39} \\
\bottomrule
\end{tabular}
}
\end{table}

\begin{figure}
\centerline{
\includegraphics[width=\linewidth, keepaspectratio]
{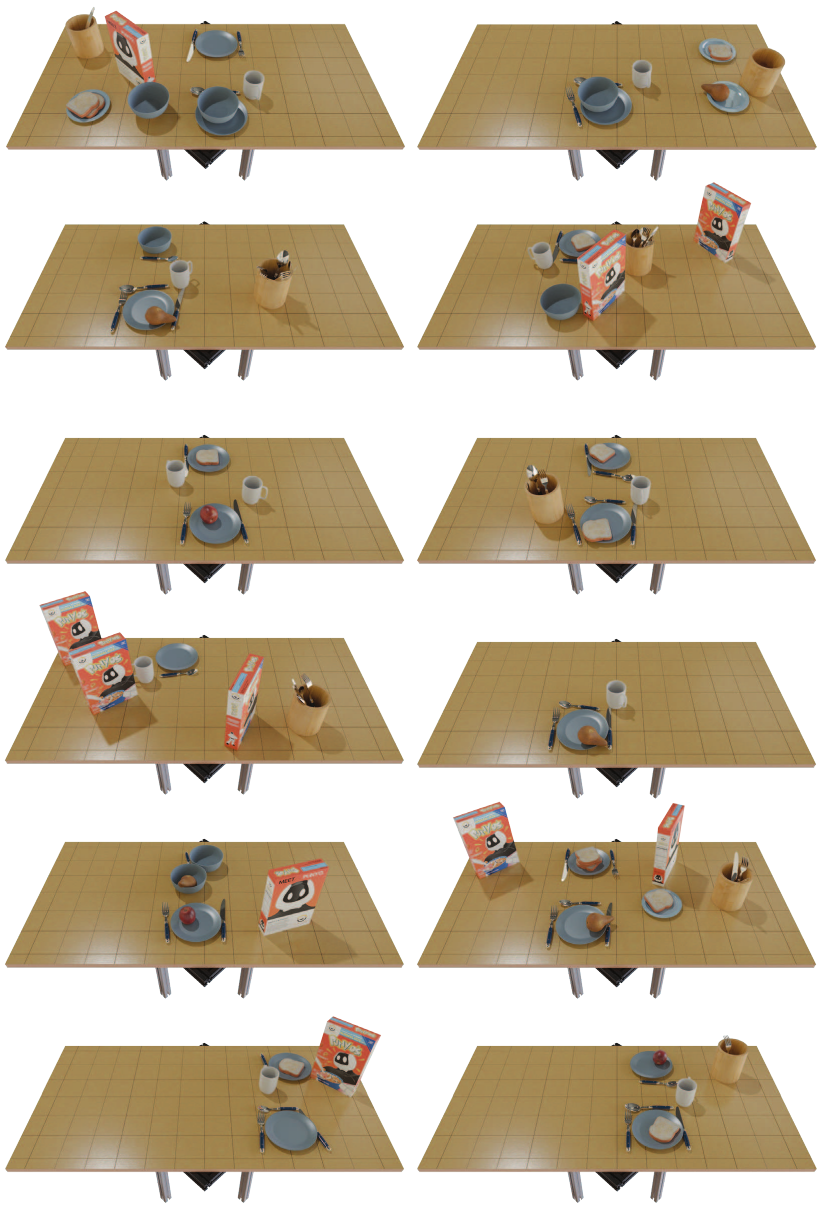}
}
\caption{Unconditional generation results for a model trained on the Breakfast Table (Low-Clutter Variant) dataset.}
\label{fig:unconditional_breakfast_low_clutter}
\end{figure}
\begin{figure}
\centerline{
\includegraphics[width=\linewidth, keepaspectratio]
{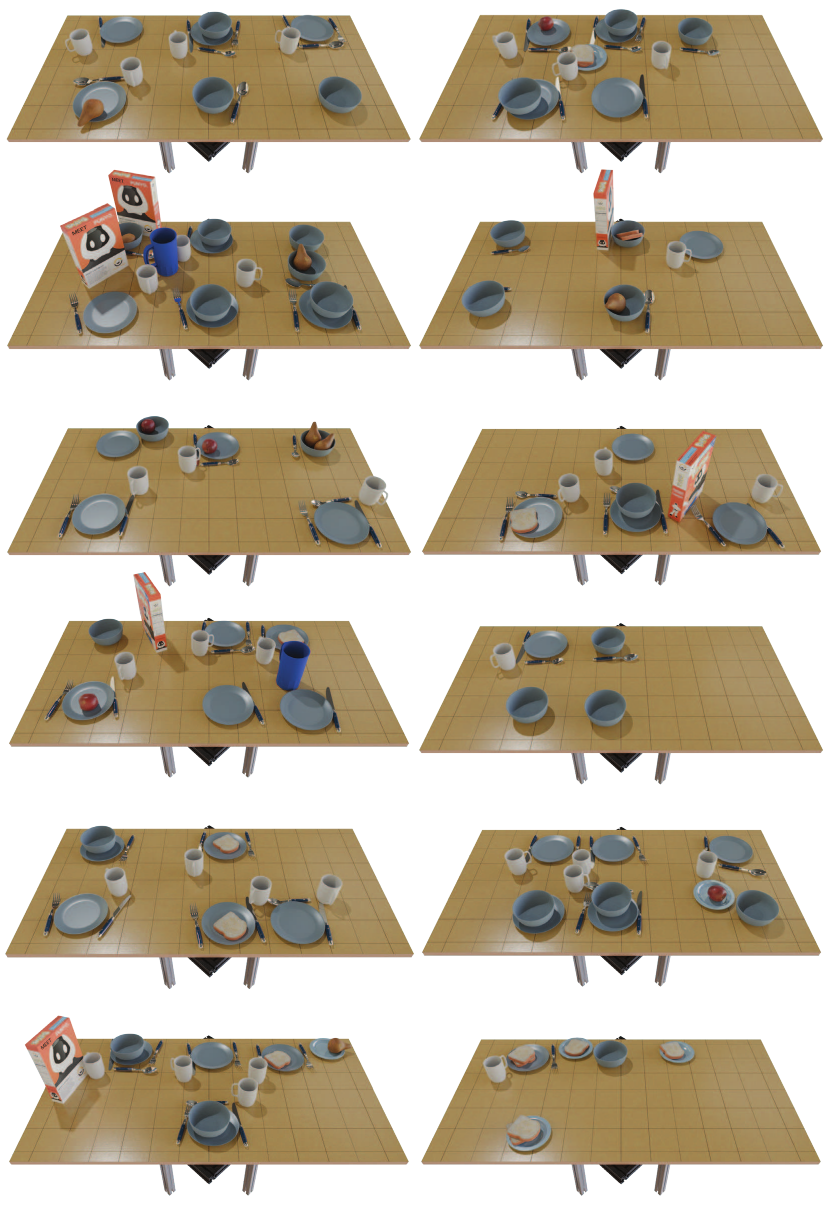}
}
\caption{Unconditional generation results for a model trained on the Breakfast Table (High-Clutter Variant) dataset.}
\label{fig:unconditional_breakfast_high_clutter}
\end{figure}
\begin{figure}
\centerline{
\includegraphics[width=\linewidth, keepaspectratio]
{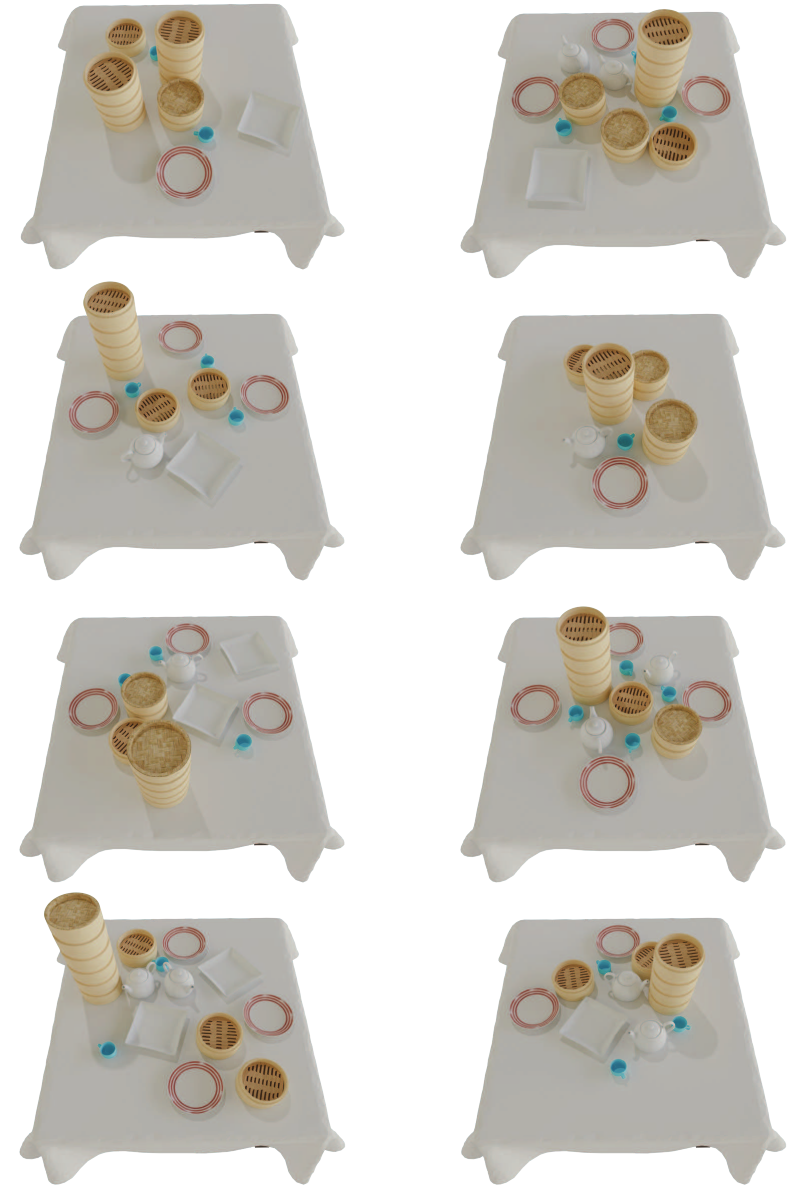}
}
\caption{Unconditional generation results for a model trained on the Dimsum Table dataset.}
\label{fig:unconditional_dimsum_table}
\end{figure}
\begin{figure}
\centerline{
\includegraphics[width=\linewidth, keepaspectratio]
{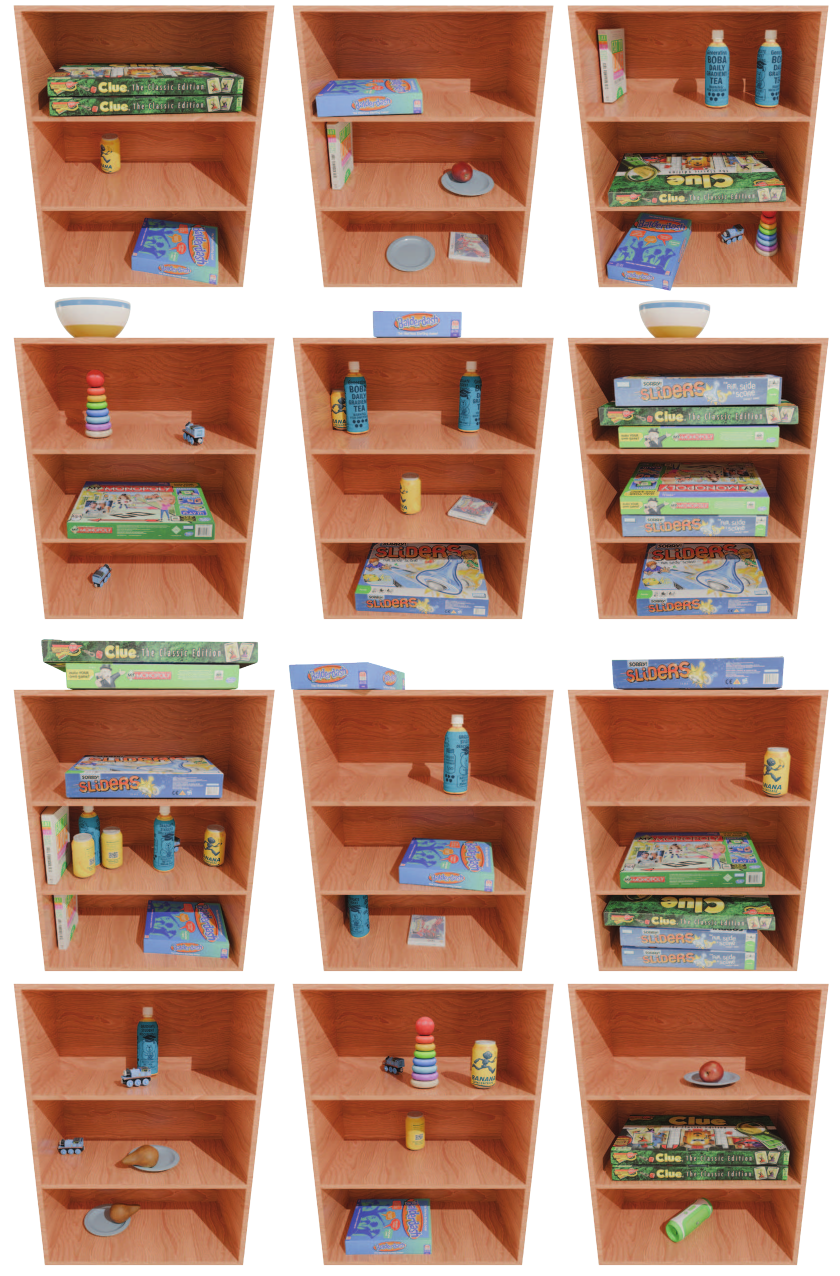}
}
\caption{Unconditional generation results for a model trained on the Living Room Shelf dataset.}
\label{fig:unconditional_living_room_shelf}
\end{figure}
\begin{figure}
\centerline{
\includegraphics[width=\linewidth, keepaspectratio]
{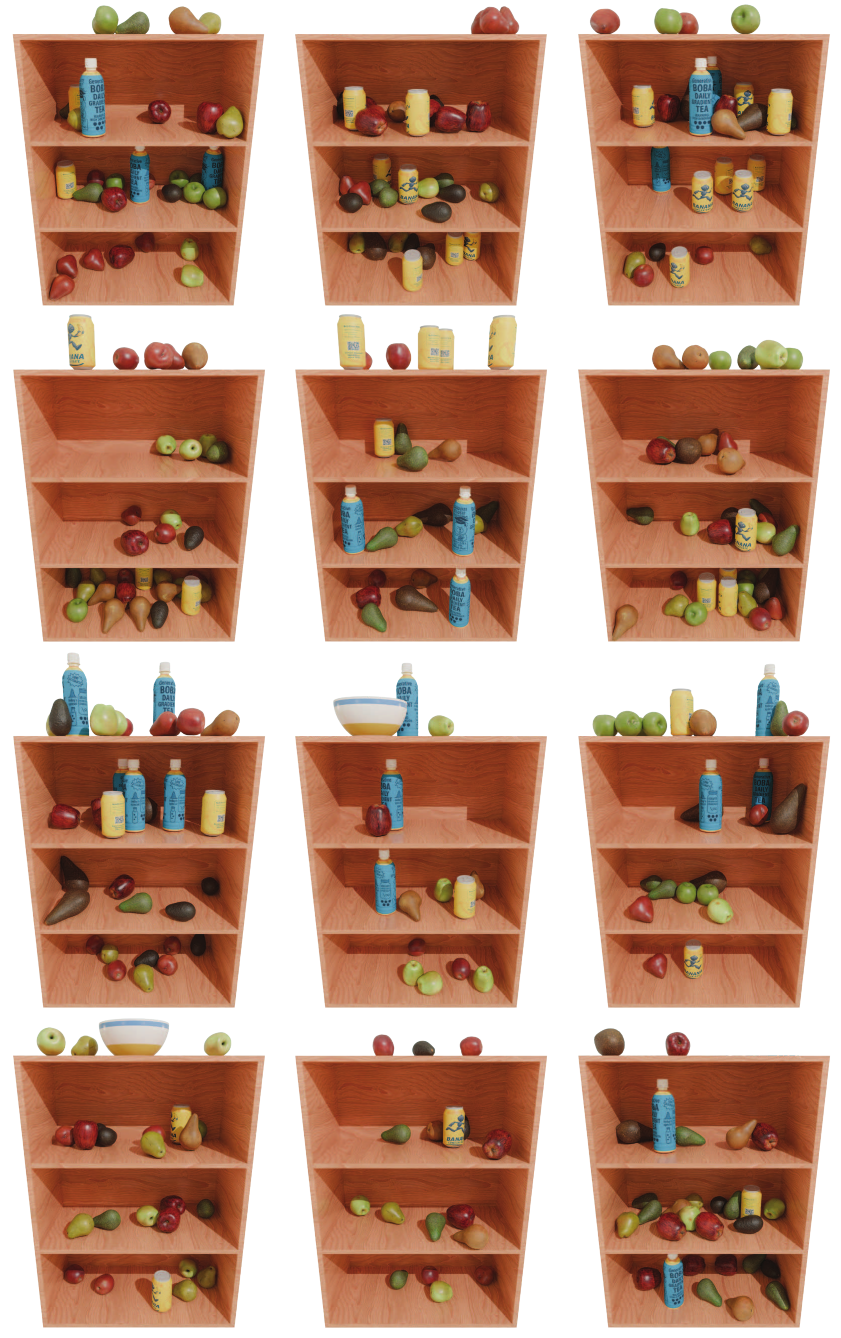}
}
\caption{Unconditional generation results for a model trained on the Pantry Shelf dataset.}
\label{fig:unconditional_pantry_shelf}
\end{figure}
\begin{figure}
\centerline{
\includegraphics[width=\linewidth, keepaspectratio]
{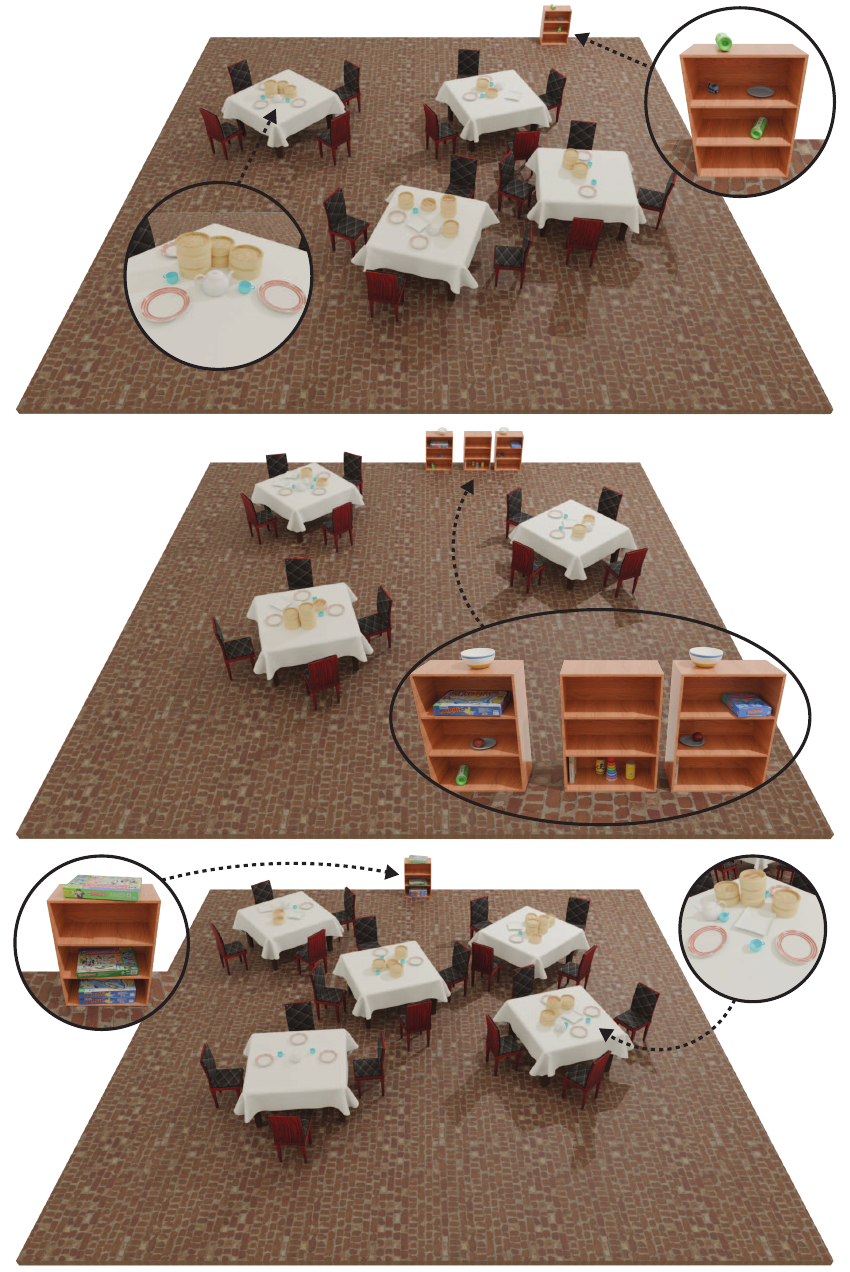}
}
\caption{Unconditional generation results for a model trained on the Restaurant (High-Clutter) dataset.}
\label{fig:unconditional_restaurant_high_clutter}
\end{figure}

\begin{figure}
\centerline{
\includegraphics[width=\linewidth, keepaspectratio]
{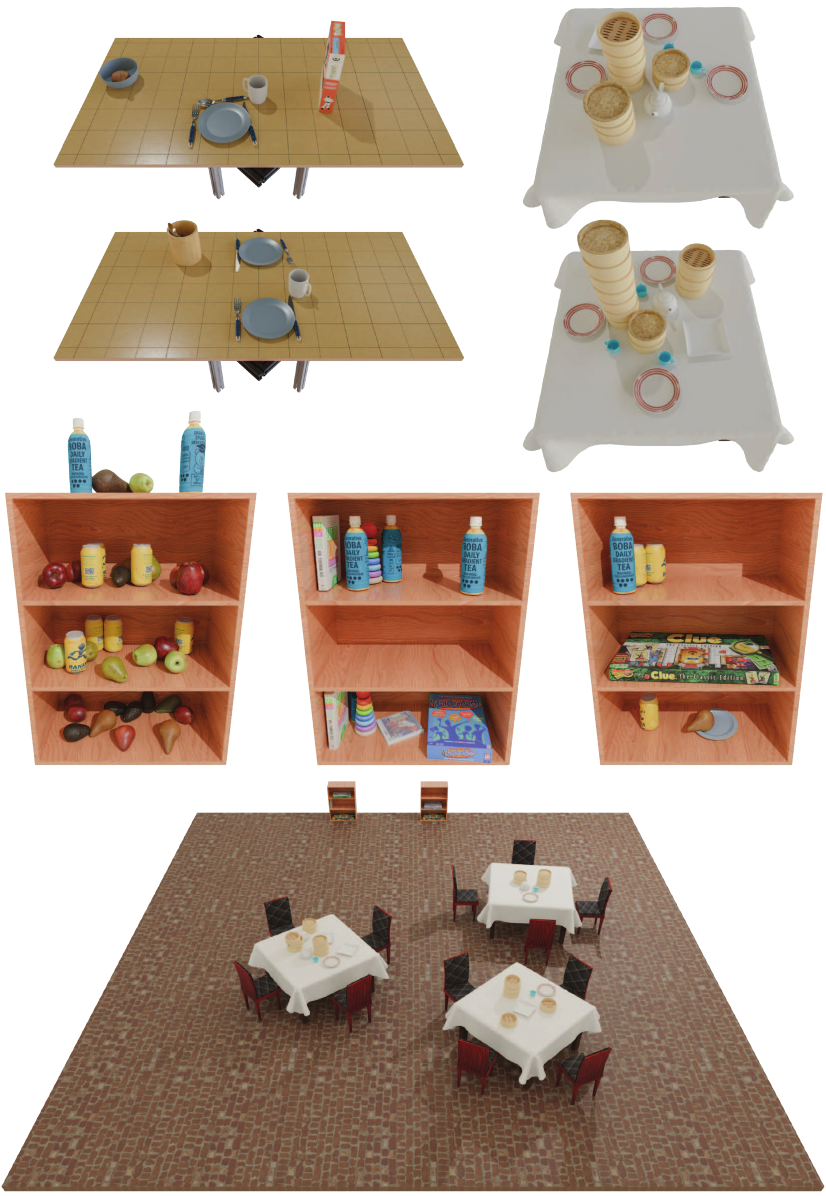}
}
\caption{Unconditional generation results from a model trained jointly on all our scenes with equal batch mixing ratios across all five scene types.}
\label{fig:unconditional_combined}
\end{figure}

We provide additional quantitative results for unconditional generation across all five datasets. Table~\ref{table:appendix-unconditional-pantry-breakfast} reports results for the Pantry Shelf and Breakfast Table (Low-Clutter) datasets, Table~\ref{table:table:appendix-unconditional-dimsum-breakfastHigh} for the Dimsum Table and Breakfast Table (High-Clutter), and Table~\ref{table:appendix-unconditional-restaurant-low} for the Restaurant (Low-Clutter) dataset.

Our model achieves the lowest classifier accuracy (CA) across all datasets, indicating strong alignment with dataset distributions. We also obtain the lowest FID on all datasets except Breakfast Table (High-Clutter), where our score is competitive and within a small margin of the best. In terms of physical realism, we achieve the lowest mean total penetration (MTP) scores on all datasets except Dimsum Table, where our score is again comparable to the best. While we do not always achieve the lowest KL divergence, all methods exhibit low KL values across datasets, suggesting it is less discriminative as an evaluation metric in this setting.

We also show qualitative unconditional generation results. Figure~\ref{fig:unconditional_breakfast_low_clutter} shows samples from a model trained on the Breakfast Table (Low-Clutter) dataset, Figure~\ref{fig:unconditional_breakfast_high_clutter} from Breakfast Table (High-Clutter), Figure~\ref{fig:unconditional_dimsum_table} from Dimsum Table, Figure~\ref{fig:unconditional_living_room_shelf} from Living Room Shelf, Figure~\ref{fig:unconditional_pantry_shelf} from Pantry Shelf, and Figure~\ref{fig:unconditional_restaurant_high_clutter} from Restaurant (High-Clutter). These visualizations provide an overview of the types of scenes in our datasets and the outputs produced by our generative model. Finally, Figure~\ref{fig:unconditional_combined} shows samples from a model trained jointly on all scene types, as described in Section~\ref{appendix:cotraining_recipe}, demonstrating that the model remains effective under multi-distribution training.

\subsection{Post Training with Reinforcement Learning}

\begin{figure}
\centerline{
\includegraphics[width=\linewidth, keepaspectratio]
{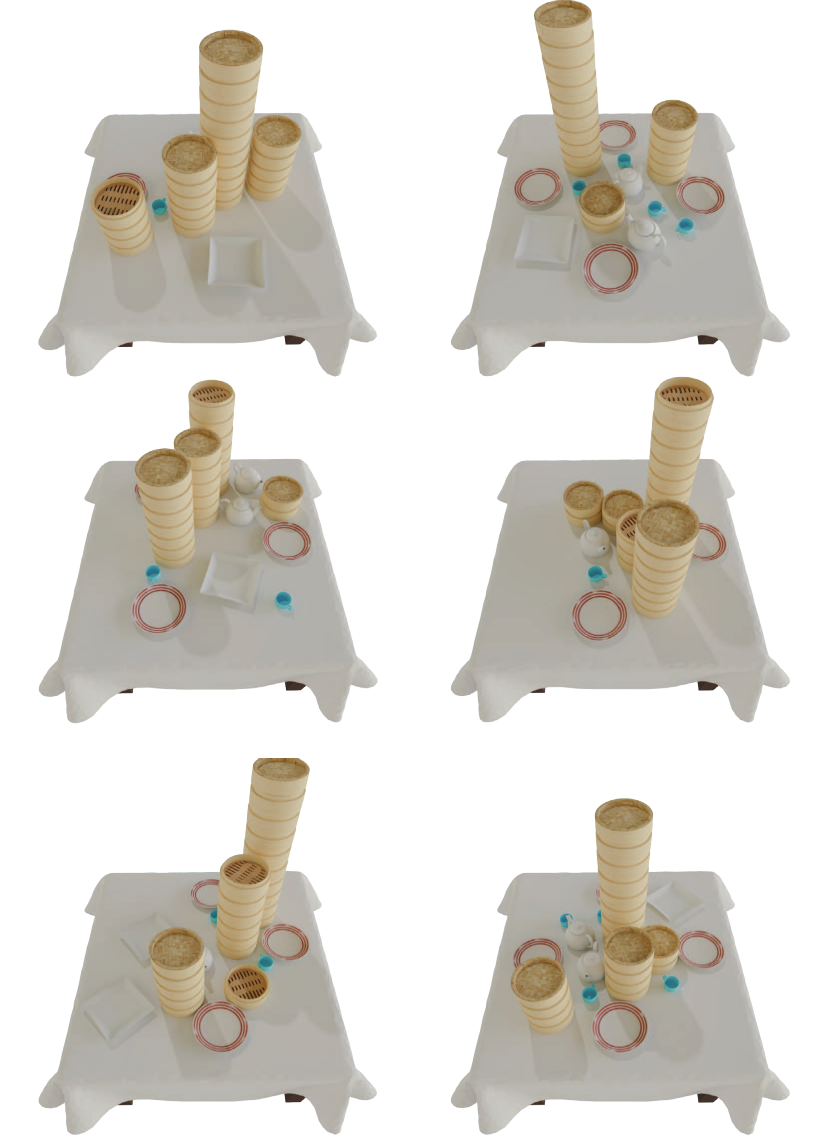}
}
\caption{Unconditional generation results from a model trained on the Dimsum dataset after post training with RL using an object number reward. Notice that the model learns to make tall steamer tray stacks.}
\label{fig:rl_dimsum}
\end{figure}
\begin{figure}
\centerline{
\includegraphics[width=\linewidth, keepaspectratio]
{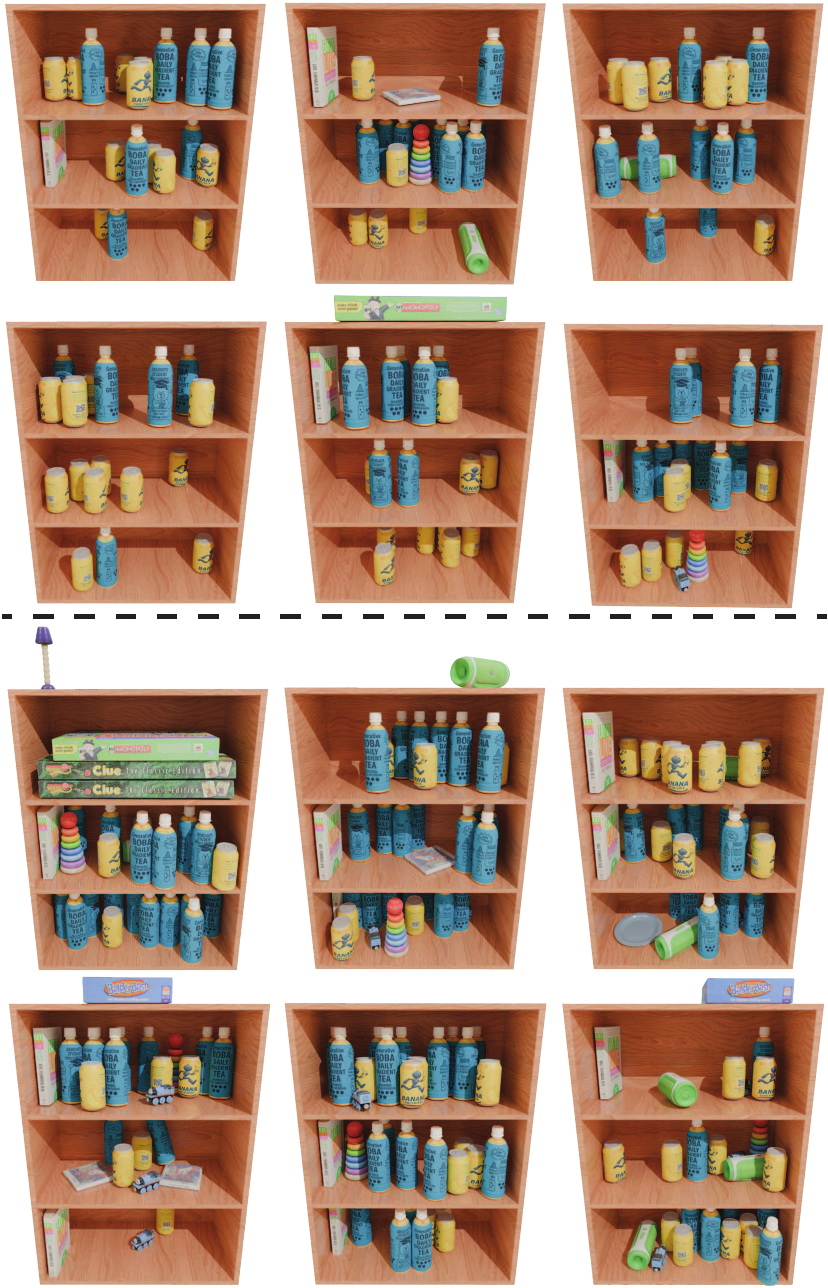}
}
\caption{Unconditional generation results from a model trained on the Living Room Shelf dataset after post training with RL using an object number reward. Samples below the dotted line are from increasing the maximum object number by 20 before post training.}
\label{fig:rl_living}
\end{figure}

We show additional unconditional generation results after post training with reinforcement learning using an object count reward. Figure~\ref{fig:rl_dimsum} displays samples from a model trained on the Dimsum Table dataset, where RL fine-tuning leads to tall steamer tray stacks that were rarely seen during pretraining. Figure~\ref{fig:rl_living} shows samples from the Living Room Shelf dataset, where we also increase the maximum object capacity by 20 prior to post training. The model adapts to fill this expanded capacity, generating denser shelf scenes. These results further support the conclusion that RL post training can effectively steer generative models toward task-aligned objectives without retraining from scratch.

\subsection{Conditional Generation}

\begin{table}[h]
\centering
\caption{Conditional generation results on the Dimsum Table and Breakfast Table (Low-Clutter) datasets. * indicates that we adjusted the methods for compatibility with our scene representation.}
\label{table:appendix-conditional}
\resizebox{\textwidth}{!}{%
\begin{tabular}{lccccccccc}
\toprule
\multirow{2}{*}{Method} 
& \multicolumn{4}{c}{Dimsum Table} 
& \multicolumn{4}{c}{Breakfast Table (Low-Clutter Variant)} \\
\cmidrule(lr){2-5} \cmidrule(lr){6-9}
& CA (100 it, \%) $\downarrow$ & KL ($\times 10^4$) $\downarrow$ & FID $\downarrow$ & APF $\uparrow$
& CA (25 it, \%) $\downarrow$ & KL ($\times 10^4$) $\downarrow$ & FID $\downarrow$ & APF $\uparrow$ \\
\midrule
DiffuScene* \cite{diffuscene}     & 87.64 ± 6.34 & 0.22  & 0.94 & \textbf{0.90} & 69.13 ± 8.71 & 3.12 & 2.62 & 0.83 \\
MiDiffusion* \cite{MiDiffusion}   & 79.26 ± 10.48 & \textbf{0.08}  & 0.92 & 0.84 & 67.91 ± 8.29  & 1.78 & 2.53 & 0.77 \\
Ours                              & \textbf{60.85 ± 6.52} & 0.18  & \textbf{0.89} & \textbf{0.90} & \textbf{58.46 ± 5.35} & \textbf{1.34} & \textbf{2.42} & \textbf{0.87} \\
\bottomrule
\end{tabular}
}
\end{table}

\begin{table}[h]
\centering
\caption{Conditional generation results on the Living Room Shelf and Restaurant (High-Clutter) datasets. * indicates that we adjusted the methods for compatibility with our scene representation.}
\label{table:appendix-conditional-living-restaurant}
\resizebox{\textwidth}{!}{%
\begin{tabular}{lccccccccc}
\toprule
\multirow{2}{*}{Method} 
& \multicolumn{4}{c}{Living Room Shelf} 
& \multicolumn{4}{c}{Restaurant (High-Clutter Variant)} \\
\cmidrule(lr){2-5} \cmidrule(lr){6-9}
& CA (100 it, \%) $\downarrow$ & KL ($\times 10^4$) $\downarrow$ & FID $\downarrow$ & APF $\uparrow$
& CA (50 it, \%) $\downarrow$ & KL ($\times 10^4$) $\downarrow$ & FID $\downarrow$ & APF $\uparrow$ \\
\midrule
DiffuScene* \cite{diffuscene}     & 69.33 ± 1.35 & 3.74 & 2.11 & \textbf{0.99} & 87.22 ± 6.73 & \textbf{0.48} & 1.42 & 0.61 \\
MiDiffusion* \cite{MiDiffusion}   & 72.66 ± 3.01 & \textbf{1.31} & 2.10 & 0.97 & 77.97 ± 11.16 & 0.66 & \textbf{1.31} & 0.53 \\
Ours                              & \textbf{54.23 ± 2.18} & 1.91 & \textbf{2.09} & \textbf{0.99} & \textbf{70.42 ± 8.19} & 0.55 & \textbf{1.31} & \textbf{0.84} \\
\bottomrule
\end{tabular}
}
\end{table}

\begin{table}[h]
\centering
\caption{Conditional generation results on the Restaurant (Low-Clutter) dataset. * indicates that we adjusted the methods for compatibility with our scene representation.}
\label{table:appendix-conditional-restaurant-low}
\resizebox{0.6\textwidth}{!}{%
\begin{tabular}{lcccc}
\toprule
\multirow{2}{*}{Method} 
& \multicolumn{4}{c}{Restaurant (Low-Clutter Variant)} \\
\cmidrule(lr){2-5}
& CA (100 it, \%) $\downarrow$ & KL ($\times 10^4$) $\downarrow$ & FID $\downarrow$ & APF $\uparrow$ \\
\midrule
DiffuScene* \cite{diffuscene}     & 79.58 ± 9.00 & \textbf{1.02} & 1.50 & 0.74 \\
MiDiffusion* \cite{MiDiffusion}   & 77.96 ± 7.24 & 1.31 & 1.45 & 0.59 \\
Ours                              & \textbf{67.39 ± 7.56} & 1.31 & \textbf{1.36} & \textbf{0.86} \\
\bottomrule
\end{tabular}
}
\end{table}

\begin{figure}
\centerline{
\includegraphics[width=\linewidth, keepaspectratio]
{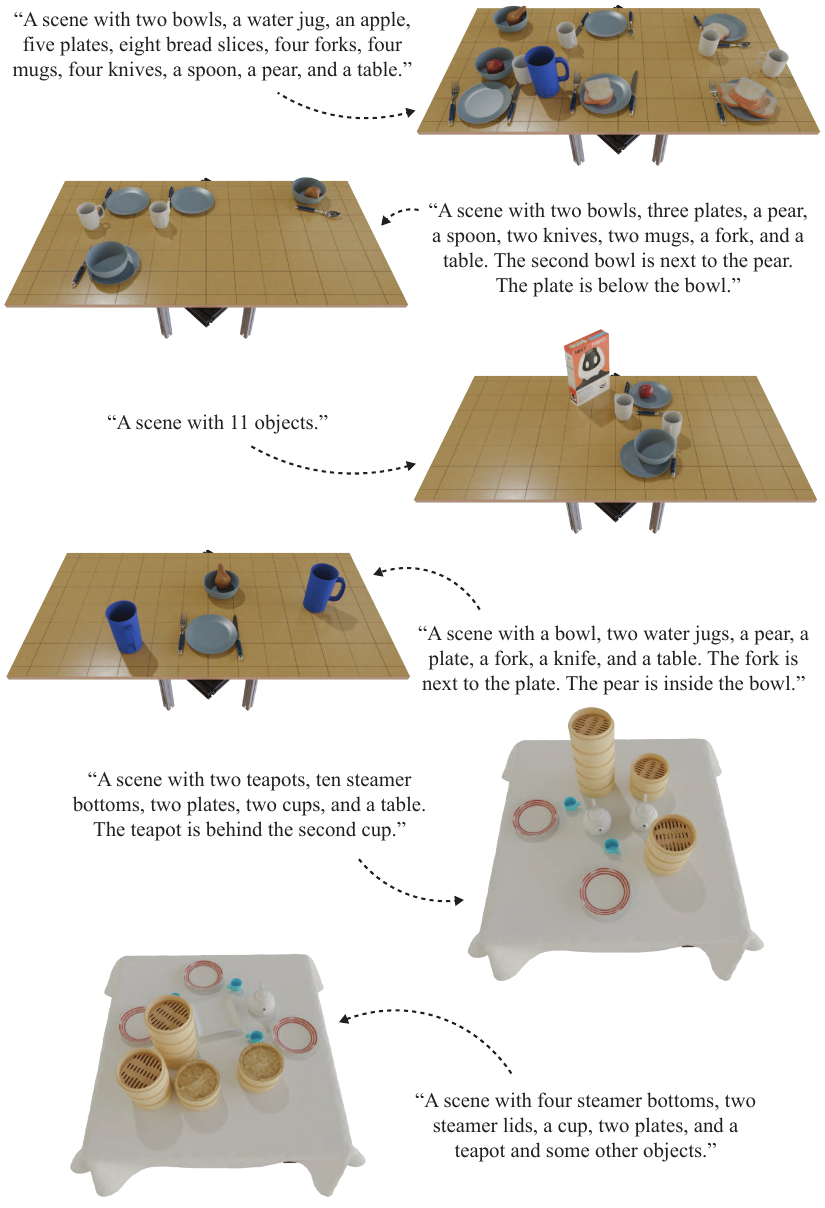}
}
\caption{Text conditional generation results for models trained on the Breakfast Table (High-Clutter), Breakfast Table (Low-Clutter), and Dimsum Table datasets.}
\label{fig:conditional_1}
\end{figure}
\begin{figure}
\centerline{
\includegraphics[width=\linewidth, keepaspectratio]
{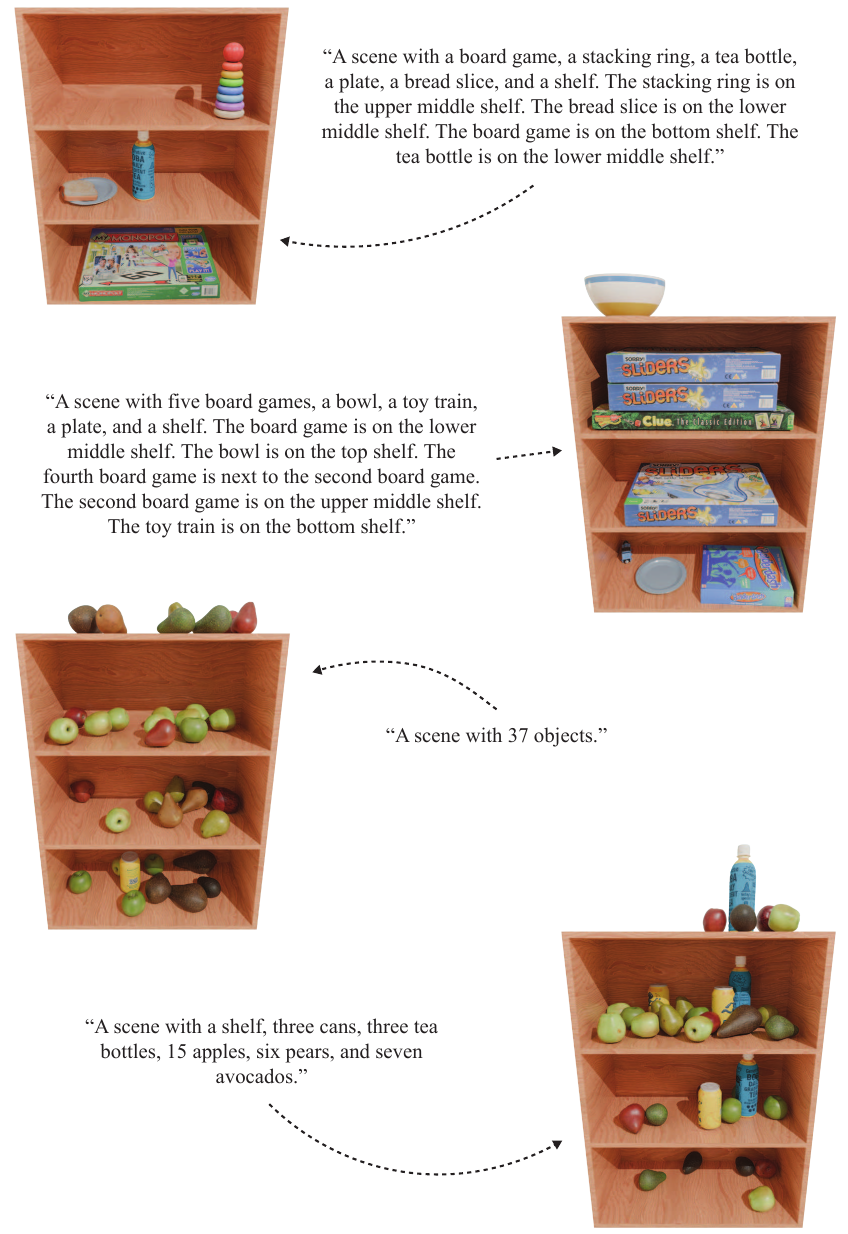}
}
\caption{Text conditional generation results for models trained on the Living Room Shelf and Pantry Shelf datasets.}
\label{fig:conditional_2}
\end{figure}
\begin{figure}
\centerline{
\includegraphics[width=\linewidth, keepaspectratio]
{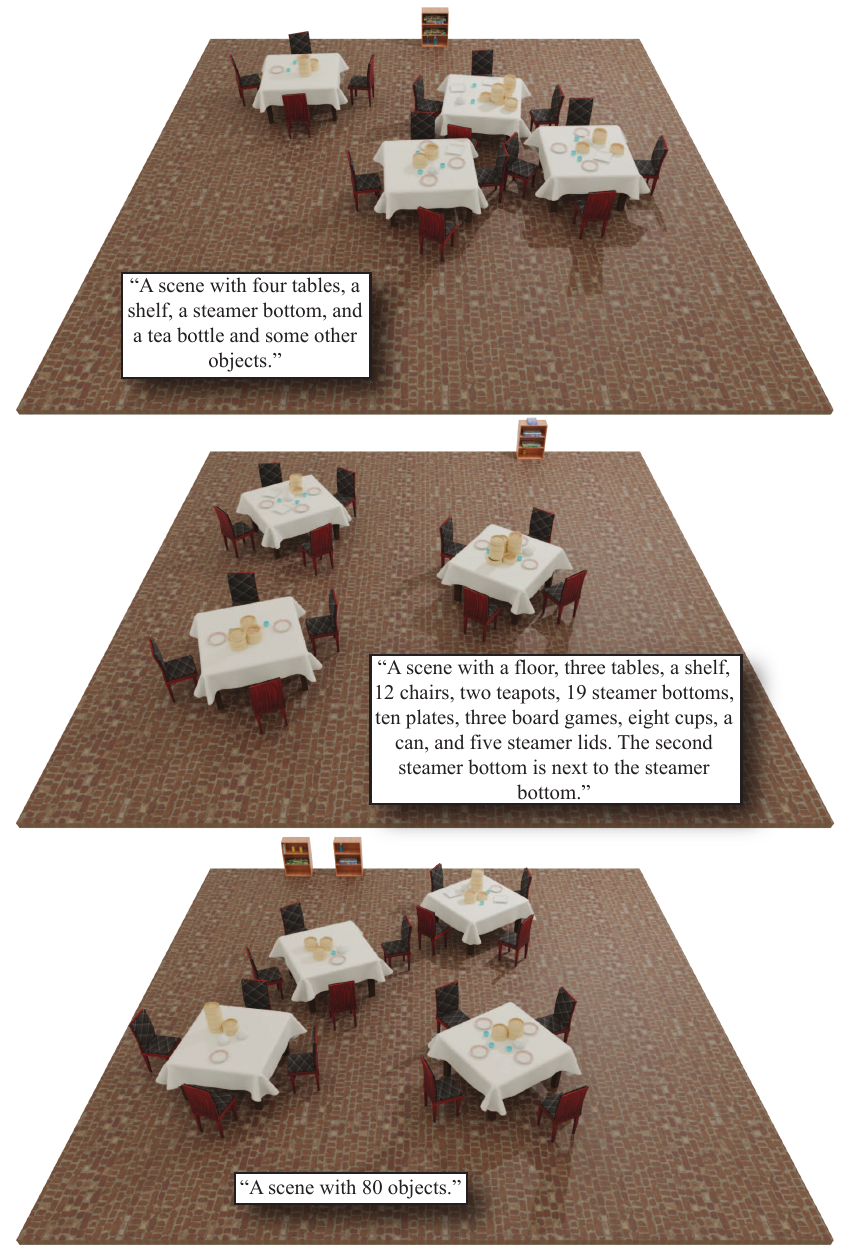}
}
\caption{Text conditional generation results for a model trained on the Restaurant (High-Clutter) dataset.}
\label{fig:conditional_3}
\end{figure}

\begin{figure}
\centerline{
\includegraphics[width=\linewidth, keepaspectratio]
{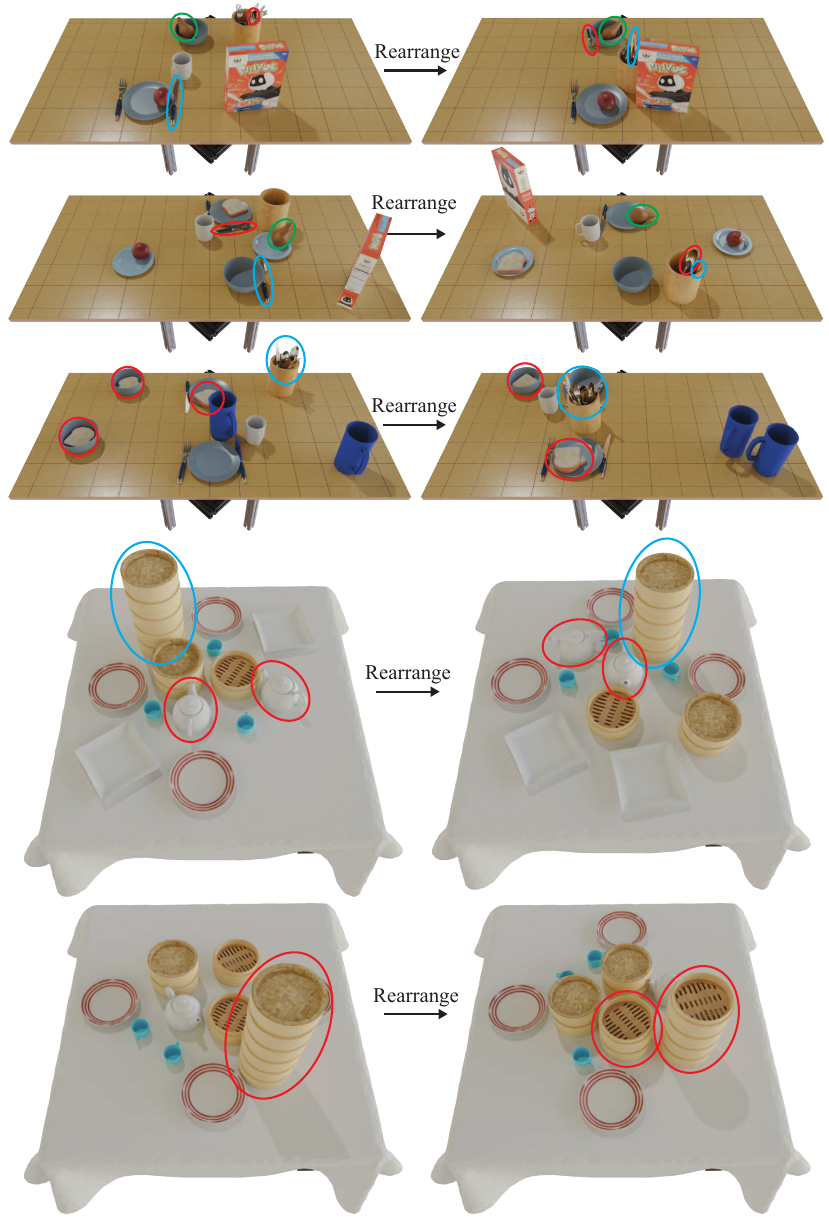}
}
\caption{Rearrangement results for models trained on the Breakfast Table and Dimsum datasets. Some objects are highlighted in red, green, and blue ellipses to facilitate easier associations before and after rearrangement.}
\label{fig:rearrangement_1}
\end{figure}
\begin{figure}
\centerline{
\includegraphics[width=\linewidth, keepaspectratio]
{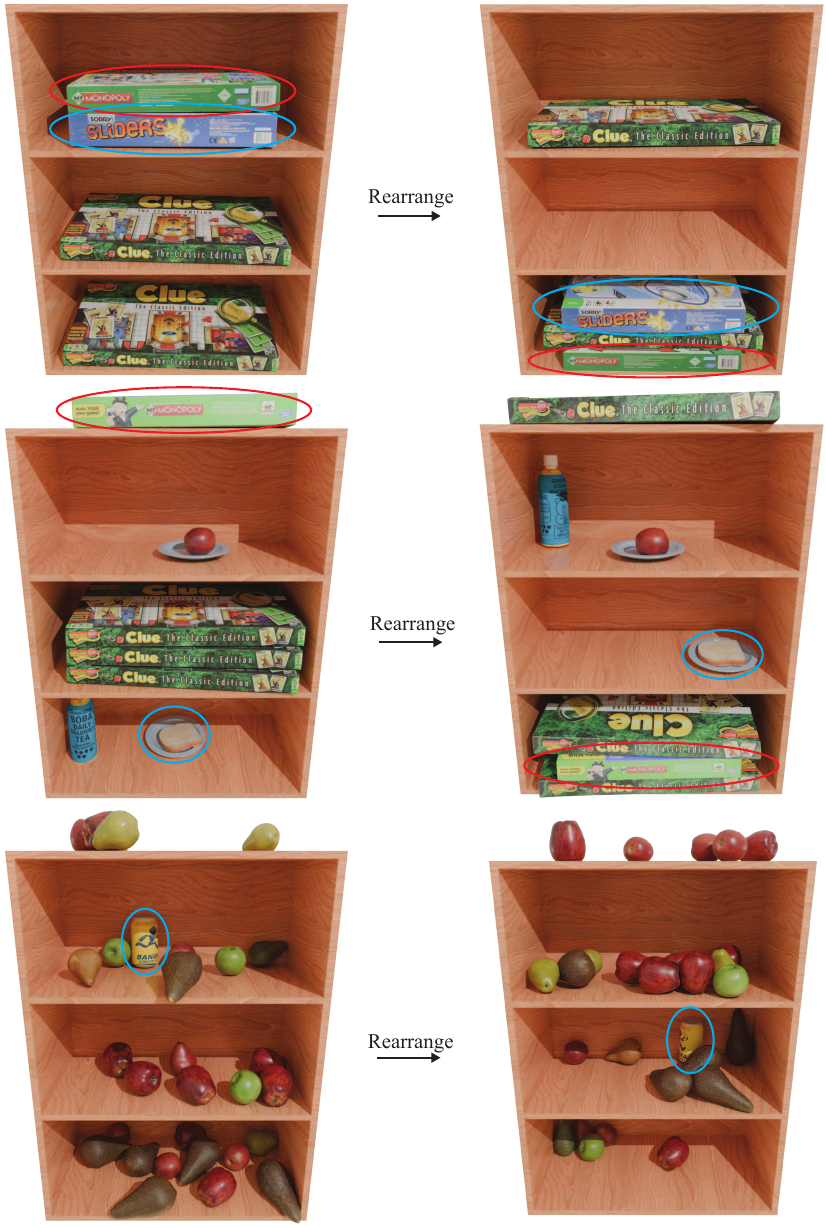}
}
\caption{Rearrangement results for models trained on the Living Room Shelf and Pantry Shelf datasets. Some objects are highlighted in red and blue ellipses to facilitate easier associations before and after rearrangement.}
\label{fig:rearrangement_2}
\end{figure}
\begin{figure}
\centerline{
\includegraphics[width=\linewidth, keepaspectratio]
{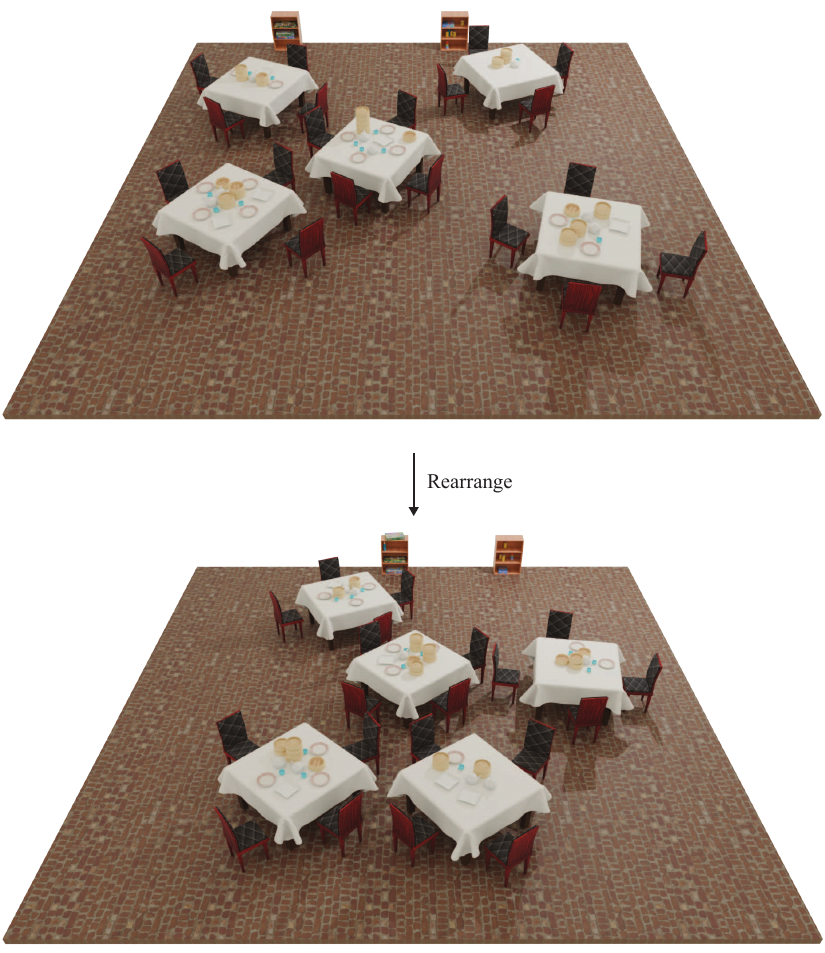}
}
\caption{Rearrangement results for a model trained on the Restaurant (High-Clutter) dataset.}
\label{fig:rearrangement_3}
\end{figure}

\begin{figure}
\centerline{
\includegraphics[width=\linewidth, keepaspectratio]
{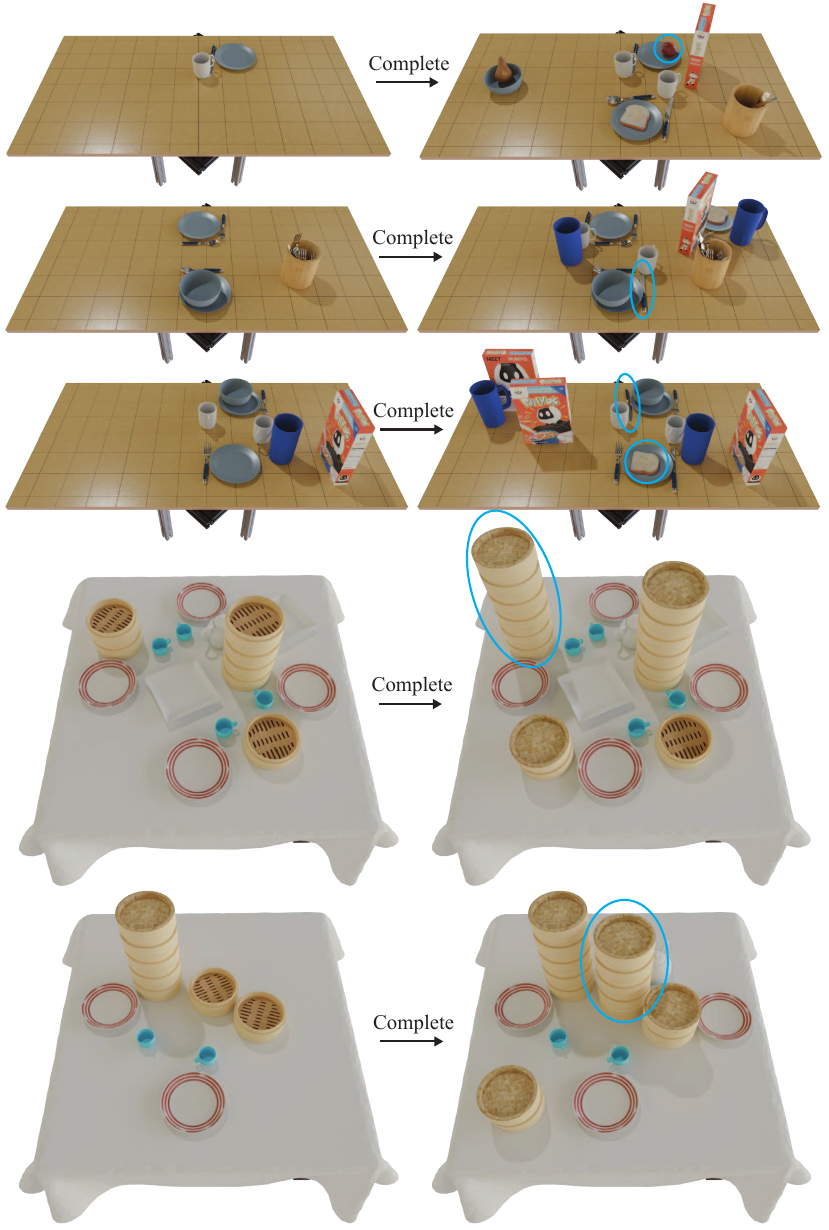}
}
\caption{Completion results for models trained on the Breakfast Table and Dimsum datasets. Some of the added objects are highlighted with blue ellipses.}
\label{fig:completion_1}
\end{figure}
\begin{figure}
\centerline{
\includegraphics[width=\linewidth, keepaspectratio]
{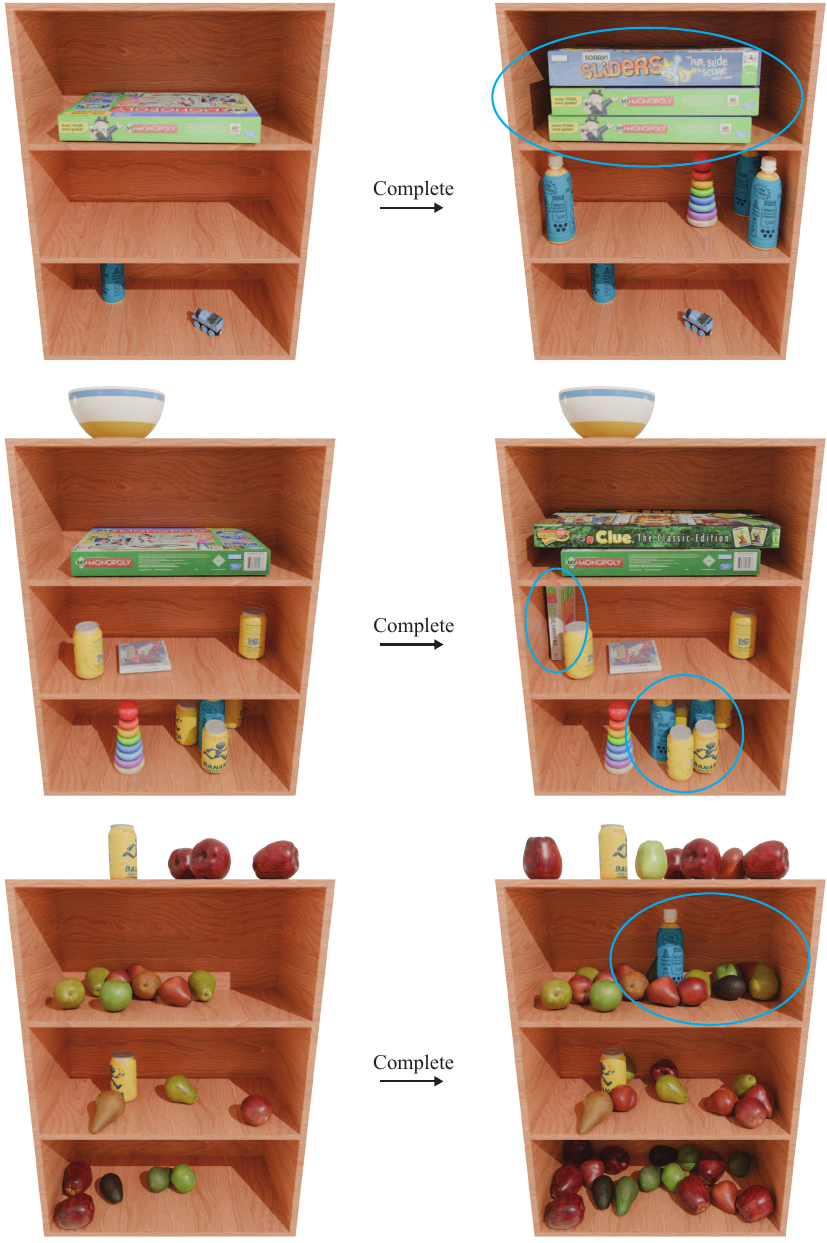}
}
\caption{Completion results for models trained on the Living Room Shelf and Pantry Shelf datasets. Some of the added objects are highlighted with blue ellipses.}
\label{fig:completion_2}
\end{figure}

\begin{figure}
\centerline{
\includegraphics[width=\linewidth, keepaspectratio]
{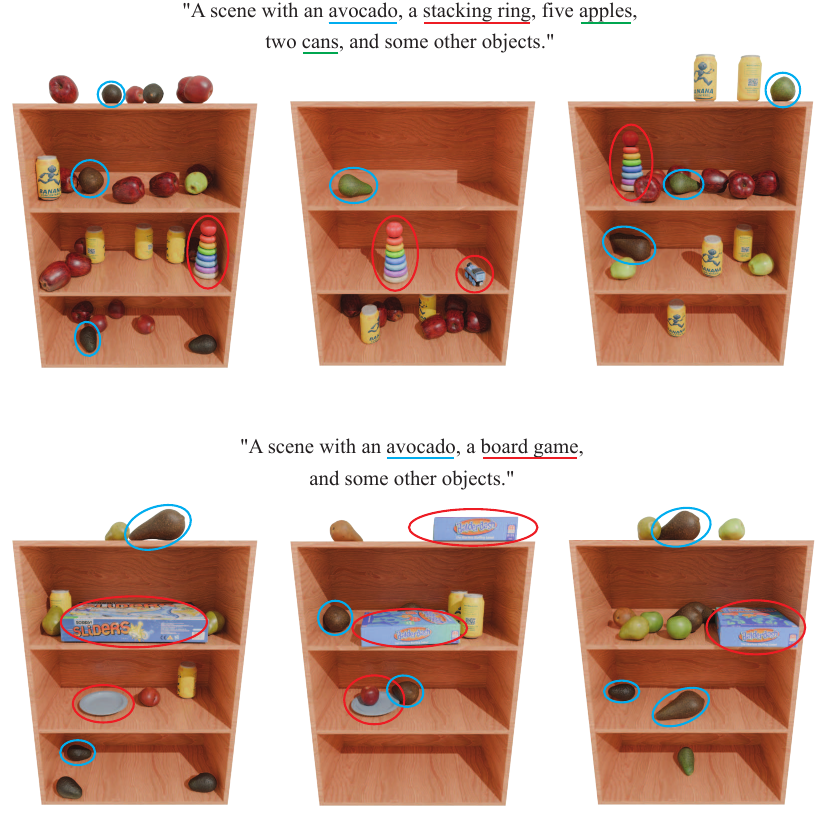}
}
\caption{Conditional generation results from a model jointly trained on the Living Room Shelf and Pantry Shelf datasets. The results show that the model can interpolate between the two data distributions when guided by text conditioning. All results use a classifier-free guidance weight of 5. Some of the objects that only appear in the Living Room Shelf dataset are highlighted in red, while some of the objects that only appear in the Pantry Shelf dataset are highlighted in blue.}
\label{fig:shelf_interpolation_appendix}
\end{figure}

We provide additional results for conditional generation across all datasets. Quantitative results are reported in Table~\ref{table:appendix-conditional} for the Dimsum Table and Breakfast Table (Low-Clutter) datasets, Table~\ref{table:appendix-conditional-living-restaurant} for the Living Room Shelf and Restaurant (High-Clutter) datasets, and Table~\ref{table:appendix-conditional-restaurant-low} for the Restaurant (Low-Clutter) dataset.

Our method achieves the lowest classifier accuracy (CA) across all datasets, indicating the strongest alignment between generated scenes and input prompts. We also obtain the highest prompt-following accuracy (APF) and the lowest FID across all datasets, along with competitive KL scores. These results suggest that our model effectively balances prompt adherence and scene realism.

We include qualitative examples for three conditional generation tasks. Prompted generation results are shown in Figures~\ref{fig:conditional_1}–\ref{fig:conditional_3}, illustrating scene synthesis from text prompts across all datasets. Rearrangement results are shown in Figures~\ref{fig:rearrangement_1}–\ref{fig:rearrangement_3}, where the model regenerates SE(3) object poses while keeping object asset IDs fixed. Completion results are shown in Figures~\ref{fig:completion_1}–\ref{fig:completion_2}, where the model fills in initially empty object slots, adding new objects while preserving the rest of the scene. Figure~\ref{fig:shelf_interpolation_appendix} demonstrates cross-dataset interpolation using a model jointly trained on Living Room Shelf and Pantry Shelf scenes, showing smooth transitions between distributions when guided by interpolated prompts.

\subsection{Inference-Time Search}

\begin{figure}
\centerline{
\includegraphics[width=\linewidth, keepaspectratio]
{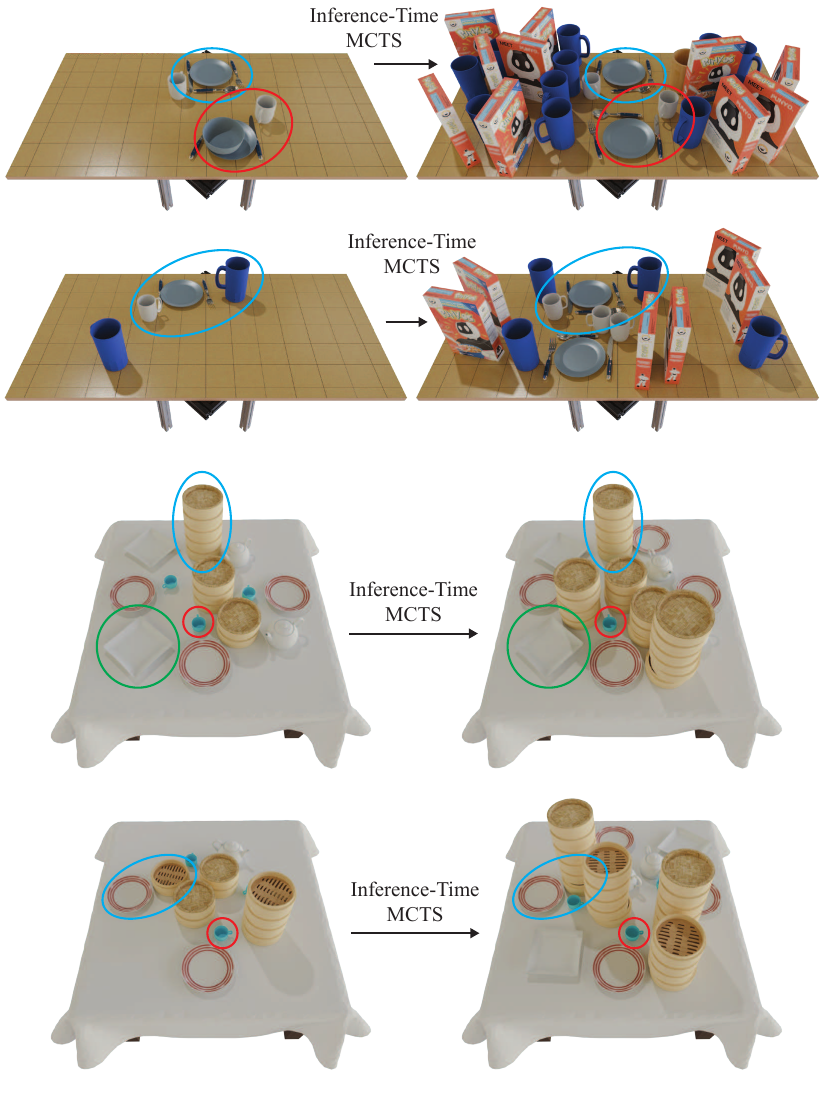}
}
\caption{Inference-time MCTS results for models trained on the Restaurant (Low-Clutter) and Dimsum datasets. Initial samples are shown on the left, and final samples (after search) on the right. Objects that remain constant are highlighted with colored ellipses.}
\label{fig:mcts_results1}
\end{figure}
\begin{figure}
\centerline{
\includegraphics[width=\linewidth, keepaspectratio]
{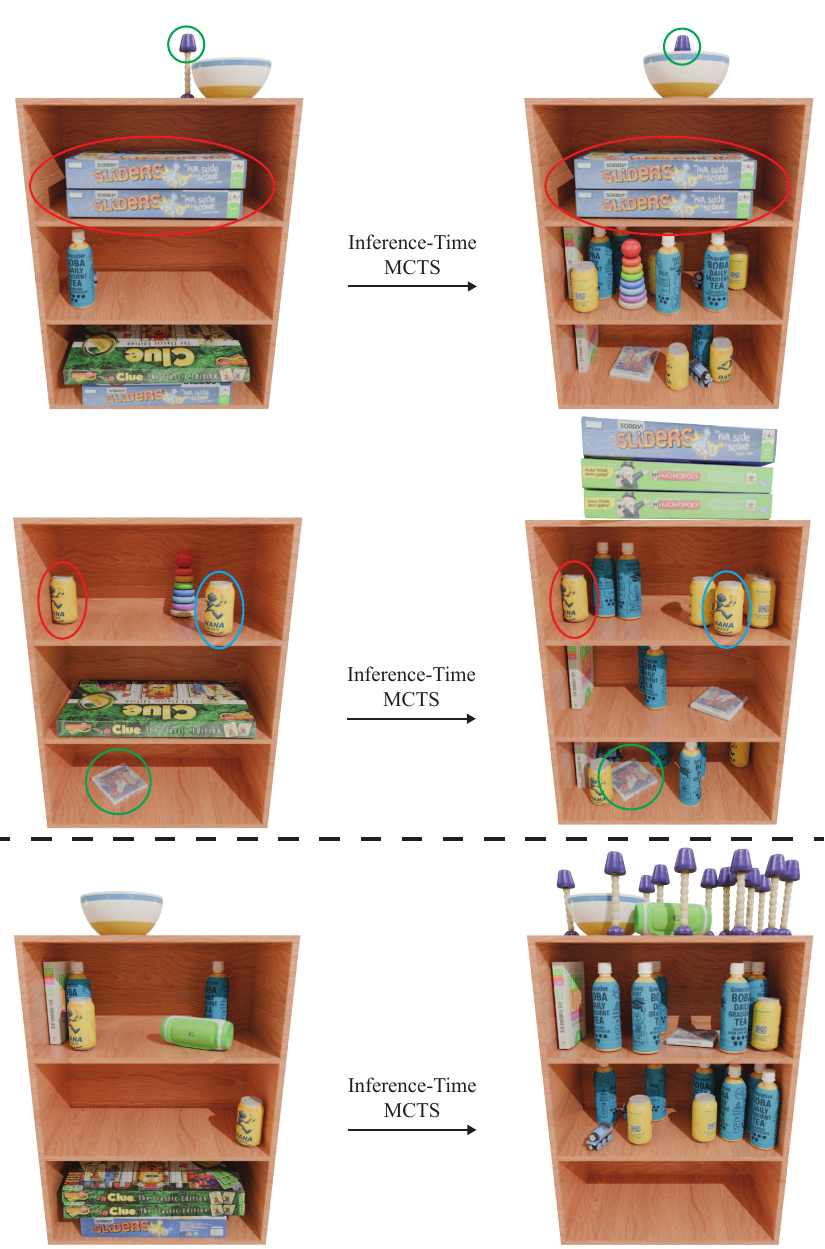}
}
\caption{Inference-time MCTS results for models trained on the Living Room Shelf dataset. Initial samples are shown on the left, and final samples (after search) on the right. Objects that remain constant are highlighted with colored ellipses. The result below the dotted line is from increasing the maximum number of objects allowed by the scene representation by 30 before the search.}
\label{fig:mcts_results2}
\end{figure}
\begin{figure}
\centerline{
\includegraphics[width=\linewidth, keepaspectratio]
{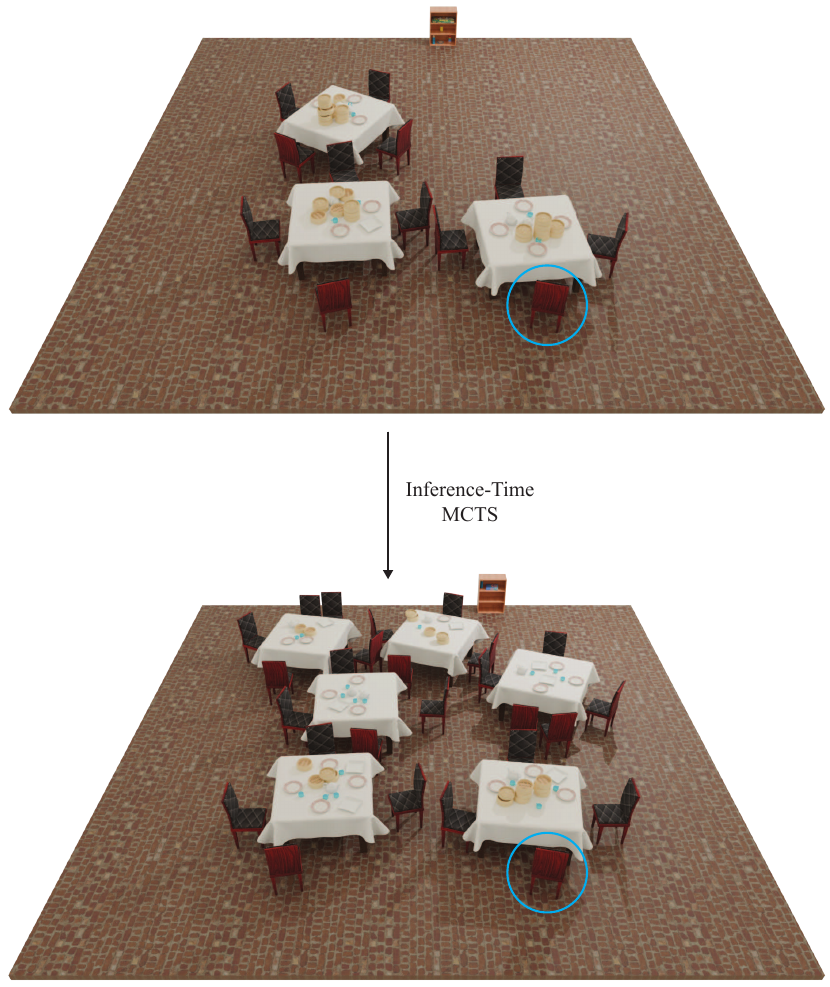}
}
\caption{Inference-time MCTS results for a model trained on the Restaurant (High-Clutter) dataset. The initial sample is shown at the top, and the final sample (after search) is shown at the bottom. Only the highlighted chair from the initial scene remains present in the final one.}
\label{fig:mcts_results3}
\end{figure}

We include additional qualitative results for inference-time search with MCTS across several datasets. Figures~\ref{fig:mcts_results1}–\ref{fig:mcts_results3} show initial and final scenes, illustrating how MCTS increases physical feasibility while preserving realistic structure.

Figure~\ref{fig:mcts_results1} presents examples from the Restaurant (Low-Clutter) and Dimsum datasets, where the search increases object count while maintaining coherent arrangements. Figure~\ref{fig:mcts_results2} shows results from the Living Room Shelf dataset, including an example where the object capacity was increased by 30 prior to search, demonstrating MCTS's ability to extrapolate to higher-density scenes than were seen during training. Finally, Figure~\ref{fig:mcts_results3} shows a case from the Restaurant (High-Clutter) dataset where nearly the entire initial scene is replaced, except for a single retained chair, highlighting MCTS’s flexibility in reshaping scenes when necessary.

These results show that MCTS can adapt scene samples to better meet downstream objectives without additional training.

\subsection{Can We Generate Novel Scenes?}

\begin{figure}
\centerline{
\includegraphics[width=\linewidth, keepaspectratio]
{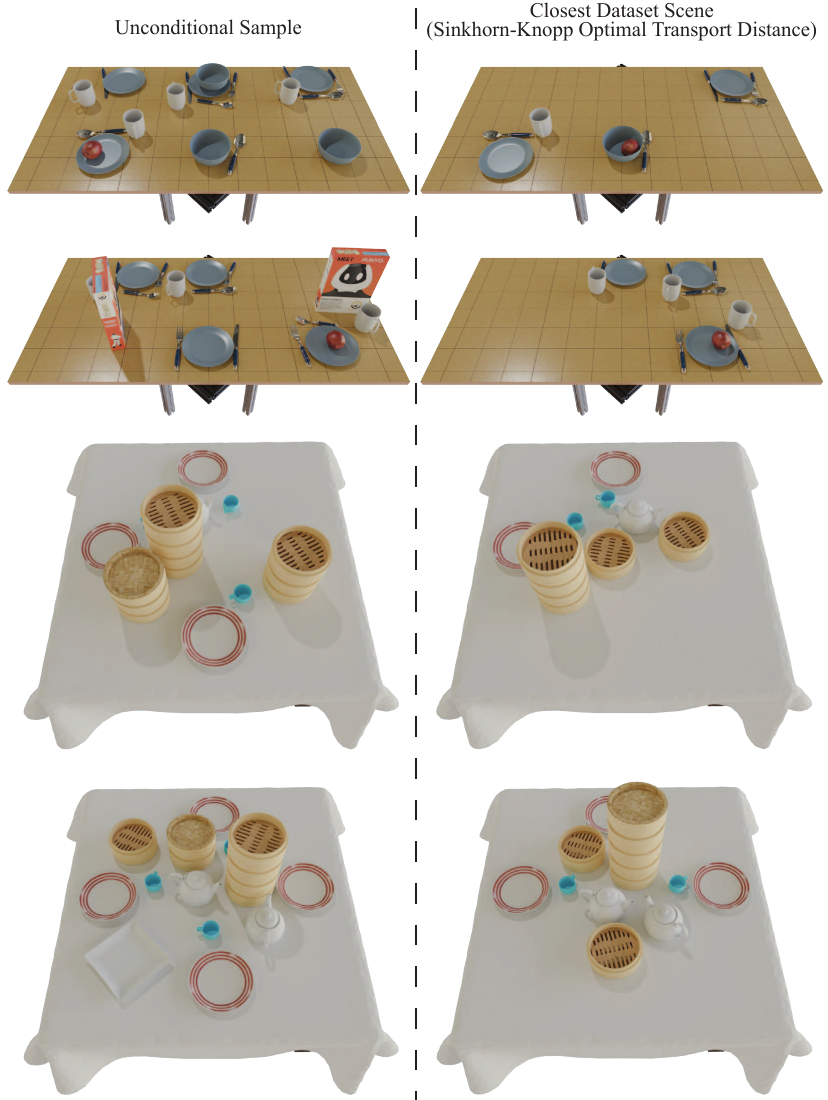}
}
\caption{Unconditional samples and their closest dataset scenes for models trained on the Breakfast Table (High-Clutter) and Dimsum Table datasets. Notice that the training scenes are quite distinct from the generated scenes, indicating that our models learn the distribution rather than memorize the training data.}
\label{fig:novel_samples1}
\end{figure}
\begin{figure}
\centerline{
\includegraphics[width=\linewidth, keepaspectratio]
{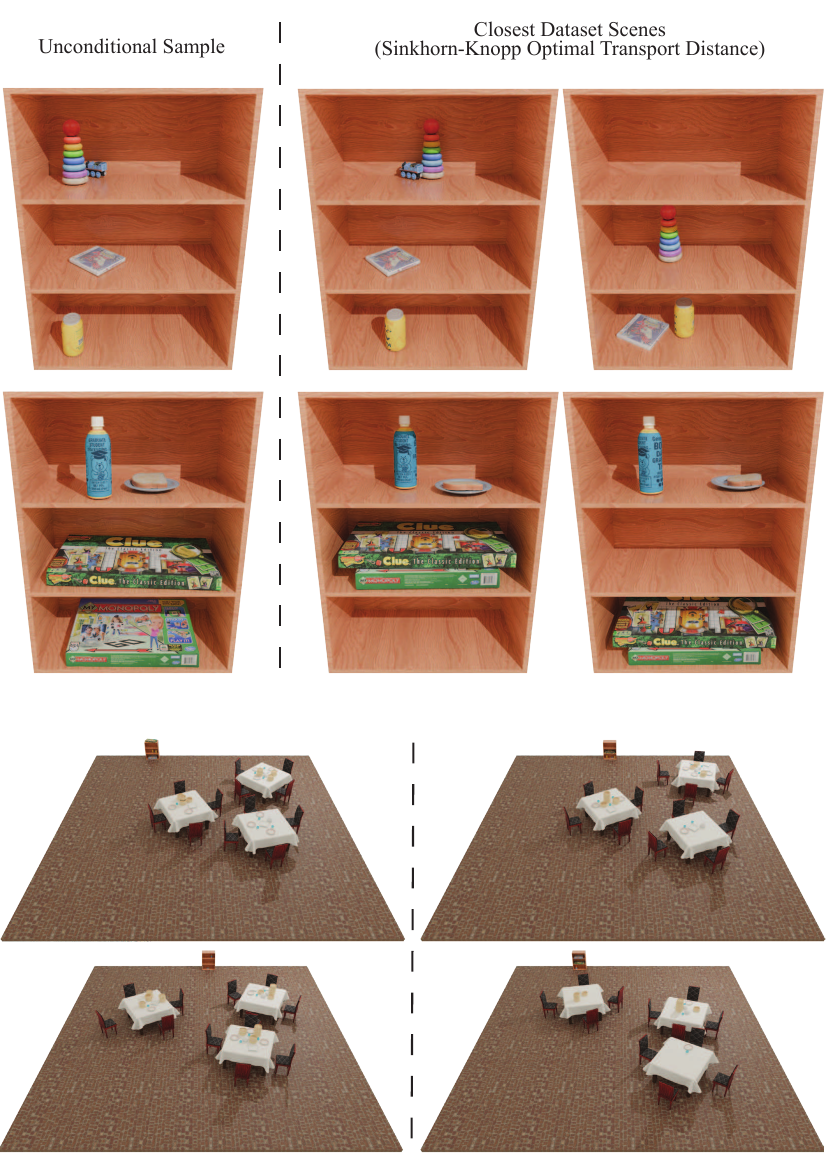}
}
\caption{Unconditional samples and their closest dataset scenes for models trained on the Living Room Shelf and Restaurant (High-Clutter) datasets. Notice that the training scenes are quite distinct from the generated scenes, indicating that our models learn the distribution rather than memorize the training data.}
\label{fig:novel_samples2}
\end{figure}

To evaluate whether our model memorizes training scenes or learns to generalize, we compare unconditional samples to their nearest neighbors in the training set using the Sinkhorn-Knopp Optimal Transport distance~\cite{silkhorn_distances_cuturi20123}. Figures~\ref{fig:novel_samples1} and~\ref{fig:novel_samples2} visualize these comparisons across four datasets. For each generated scene (left), we show its closest training scene (right).

We adopt Optimal Transport (OT) because our scene representation is an unordered set of objects, making standard vector-space distances such as L2 ill-suited. OT enables comparison between such sets while respecting permutation invariance. We utilize the Sinkhorn approximation for computational efficiency, enabling scalable pairwise comparisons between high-dimensional object sets without requiring exact matching.

Across all examples, we observe clear differences between the generated and nearest training scenes. While some similarities may exist, such as object types or broad spatial patterns, the generated scenes contain novel combinations of objects and distinct SE(3) configurations. This indicates that our model synthesizes new samples from the learned distribution rather than memorizing the training data.

These results support the conclusion that diffusion-based scene models can generate diverse, realistic, and physically plausible scenes without overfitting to the training set.

\section{Ablations}
\label{appendix:ablations}

We ablate key components of our scene generation method to understand their contribution to the final model performance. Specifically, we study the effects of the training objective, rotation representation, training precision, and post-processing pipeline.

\subsection{Training Objective}

\begin{table}[h]
\centering
\caption{Training objective ablation. Unconditional generation results on Breakfast Table (Low-Clutter) and Dimsum Table. ``Mixed`` refers to our standard model, and ``DDPM`` refers to our model trained with the continuous-only objective from \cite{diffuscene}.}
\label{table:appendix-unconditional-training-objective-breakfast-dimsum}
\resizebox{\textwidth}{!}{%
\begin{tabular}{lccccccccc}
\toprule
\multirow{2}{*}{Method} 
& \multicolumn{4}{c}{Breakfast Table (Low-Clutter Version)} 
& \multicolumn{4}{c}{Dimsum Table} \\
\cmidrule(lr){2-5} \cmidrule(lr){6-9}
& CA (25 it, \%) $\downarrow$ & KL ($\times 10^4$) $\downarrow$ & FID $\downarrow$ & MTP (cm) $\downarrow$
& CA (100 it, \%) $\downarrow$ & KL ($\times 10^4$) $\downarrow$ & FID $\downarrow$ & MTP (cm) $\downarrow$ \\
\midrule
DDPM     & 66.94 ± 8.19 & 2.16  & 2.45 & \textbf{3.33} & 76.70 ± 11.86 & \textbf{0.49} & 0.94 & \textbf{0.19} \\
Mixed   & \textbf{57.85 ± 4.98} & \textbf{1.67}  & \textbf{2.39} & 3.50  & \textbf{57.66 ± 4.70} & 0.63  & \textbf{0.89} & 0.20 \\
\bottomrule
\end{tabular}
}
\end{table}

\begin{table}[h]
\centering
\caption{Training objective ablation. Unconditional generation results on Living Room Shelf and Restaurant (High-Clutter). ``Mixed`` refers to our standard model, and ``DDPM`` refers to our model trained with the continuous-only objective from \cite{diffuscene}.}
\label{table:appendix-unconditional-training-objective-living-restaurant}
\resizebox{\textwidth}{!}{%
\begin{tabular}{lccccccccc}
\toprule
\multirow{2}{*}{Method} 
& \multicolumn{4}{c}{Living Room Shelf} 
& \multicolumn{4}{c}{Restaurant (High-Clutter Variant)} \\
\cmidrule(lr){2-5} \cmidrule(lr){6-9}
& CA (25 it, \%) $\downarrow$ & KL ($\times 10^4$) $\downarrow$ & FID $\downarrow$ & MTP (cm) $\downarrow$
& CA (50 it, \%) $\downarrow$ & KL ($\times 10^4$) $\downarrow$ & FID $\downarrow$ & MTP (cm) $\downarrow$ \\
\midrule
DDPM     & 68.48 ± 1.65 & 3.96  & 2.13 & 0.03 & 80.75 ± 5.54 & 0.88 & 1.42 & 13.59 \\
Mixed   & \textbf{52.84 ± 1.26} & \textbf{2.13} & \textbf{2.09} & \textbf{0.02}  & \textbf{70.74 ± 8.05} & \textbf{0.87}  & \textbf{1.31} & \textbf{6.31} \\
\bottomrule
\end{tabular}
}
\end{table}

We compare our mixed discrete-continuous training objective to a continuous-only DDPM objective that treats both object categories and poses as continuous variables, following \citet{diffuscene}. As shown in Tables~\ref{table:appendix-unconditional-training-objective-breakfast-dimsum} and \ref{table:appendix-unconditional-training-objective-living-restaurant}, the mixed objective consistently achieves lower classifier accuracy (CA) and FID across all datasets, indicating improved alignment with the data distribution and higher sample quality. While the DDPM-only variant slightly reduces mean total penetration (MTP) on some datasets, it performs worse on key alignment and realism metrics. These results highlight the benefit of jointly modeling discrete and continuous factors when training SE(3) scene generative models.

\subsection{Rotation Representation}

\begin{table}[h]
\centering
\caption{Rotation representation ablation. Unconditional generation results on Breakfast Table (Low Clutter) and Living Room Shelf. $\mathbb{R}^9$+SVD refers to the rotation representation from \cite{hitchhiker_guide_so3} that is used by our method.}
\label{table:appendix-unconditional-rotation-representation}
\resizebox{\textwidth}{!}{%
\begin{tabular}{lccccccccc}
\toprule
\multirow{2}{*}{Method} 
& \multicolumn{4}{c}{Breakfast Table (Low-Clutter Variant)} 
& \multicolumn{4}{c}{Living Room Shelf} \\
\cmidrule(lr){2-5} \cmidrule(lr){6-9}
& CA (25 it, \%) $\downarrow$ & KL ($\times 10^4$) $\downarrow$ & FID $\downarrow$ & MTP (cm) $\downarrow$
& CA (100 it, \%) $\downarrow$ & KL ($\times 10^4$) $\downarrow$ & FID $\downarrow$ & MTP (cm) $\downarrow$ \\
\midrule
Axis-Angle     & 66.57 ± 6.97 & 1.25 & 2.51 & \textbf{3.25} & 69.91 ± 3.90 & 2.61 & 2.20 & 0.03 \\
Quaternion     & 61.38 ± 7.64 & \textbf{1.00} & 2.46 & 3.31 & 61.58 ± 4.61 & 2.53 & 2.10 & 0.03 \\
$\mathbb{R}^9$+SVD   & \textbf{57.85 ± 4.98} & 1.67 & \textbf{2.39} & 3.50  & \textbf{52.84 ± 1.26} & \textbf{2.13} & \textbf{2.09} & \textbf{0.02} \\
\bottomrule
\end{tabular}
}
\end{table}

We evaluate three rotation parameterizations: axis-angle, quaternion, and the $\mathbb{R}^9$ representation projected onto SO(3) using SVD, following \citet{hitchhiker_guide_so3}, which is used in our method. As shown in Table~\ref{table:appendix-unconditional-rotation-representation}, the $\mathbb{R}^9$+SVD representation achieves the best classifier accuracy (CA) and FID on both datasets, with competitive mean total penetration (MTP). These results suggest that learning in an unconstrained higher-dimensional space and projecting to SO(3) at sample time yields more accurate and realistic generations.

\subsection{Training Precision}

\begin{table}[h]
\centering
\caption{Training precision ablation. Unconditional generation results on Restaurant (High-Clutter) and Living Room Shelf. ``ft32`` uses full precision (32-bit) with \texttt{matmul\_precision="high"}, while ``bfloat`` uses mixed precision (bf16) with \texttt{matmul\_precision="medium"}. Evaluation is conducted in the training precision.}
\label{table:appendix-unconditional-precision}
\resizebox{\textwidth}{!}{%
\begin{tabular}{lcccccccc}
\toprule
\multirow{2}{*}{Precision} 
& \multicolumn{4}{c}{Restaurant (High-Clutter)} 
& \multicolumn{4}{c}{Living Room Shelf} \\
\cmidrule(lr){2-5} \cmidrule(lr){6-9}
& CA (50 it, \%) $\downarrow$ & KL ($\times 10^4$) $\downarrow$ & FID $\downarrow$ & MTP (cm) $\downarrow$ 
& CA (100 it, \%) $\downarrow$ & KL ($\times 10^4$) $\downarrow$ & FID $\downarrow$ & MTP (cm) $\downarrow$ \\
\midrule
bfloat & 90.32 ± 1.68 & \textbf{0.49} & 1.65 & 9.44 & 88.46 ± 0.35 & 2.21 & 2.18 & 0.11 \\
ft32   & \textbf{70.74 ± 8.05} & 0.87 & \textbf{1.31} & \textbf{6.31} & \textbf{52.84 ± 1.26} & \textbf{2.13} & \textbf{2.10} & \textbf{0.02} \\
\bottomrule
\end{tabular}
}
\end{table}

We compare models trained with full precision (32-bit) and mixed precision (bfloat16), evaluating each using the same precision setting as used during training. As shown in Table~\ref{table:appendix-unconditional-precision}, full precision consistently outperforms mixed precision across CA, FID, and MTP, with especially large gains on MTP, reflecting the metric’s sensitivity to small physical violations near the boundary between static equilibrium and penetration. These results suggest that while mixed precision may reduce training cost and is useful during development, it can degrade generation quality for physically grounded tasks that require fine-grained accuracy.

\subsection{Post Processing Pipeline}

\begin{figure}
\centerline{
\includegraphics[width=\linewidth, keepaspectratio]
{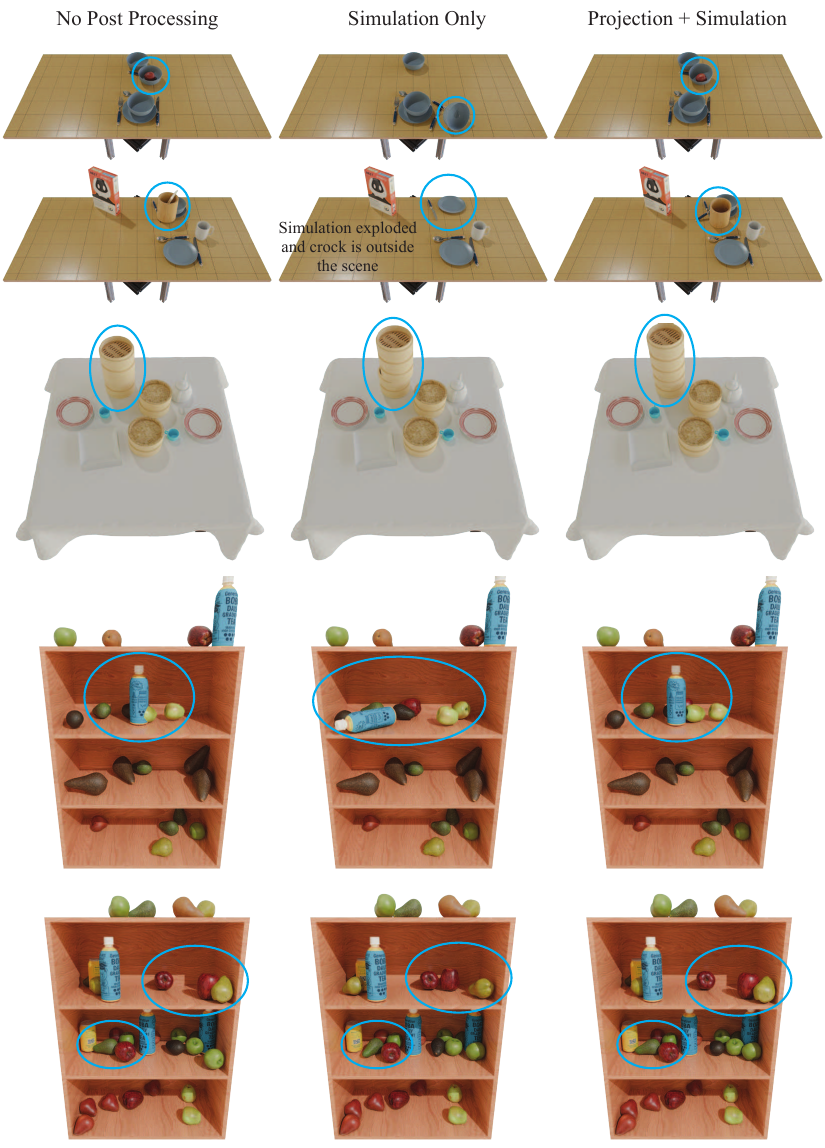}
}
\caption{\textbf{Qualitative post processing ablation.} Interesting regions are highlighted with blue ellipses. The left column contains samples generated by the model without post processing. The middle column contains samples from the left after applying physics simulation only. The column on the right contains samples from the left after applying our complete post processing (non-penetration projection followed by physics simulation). Note that samples in the left column are not physically feasible. Samples in the other columns are physically feasible, but the samples in the right column are closer to the desired scenes from the left column.}
\label{fig:post_processing_ablation}
\end{figure}

We qualitatively ablate the effects of our post-processing pipeline in Figure~\ref{fig:post_processing_ablation}, comparing three variants: (1) no post-processing, (2) simulation only, and (3) our full pipeline with projection followed by simulation. The left column shows that unprocessed samples often contain physical violations, such as interpenetrations or unsupported objects. The middle column demonstrates that simulation alone can resolve some issues, but may also destabilize the scene. In contrast, our full pipeline (right column) eliminates interpenetrations while preserving the intended structure, resulting in physically stable and realistic scenes. These results highlight the importance of combining geometric projection and physics simulation to ensure feasibility. To make the comparison informative, we selected examples with prominent initial physical violations.

\end{document}